\documentclass{article}
\usepackage{kr}

\usepackage{times}
\usepackage{soul}
\usepackage{url}
\usepackage[hidelinks]{hyperref}
\usepackage[utf8]{inputenc}
\usepackage[small]{caption}
\usepackage[table]{xcolor}
\usepackage{graphicx}
\usepackage{amsmath}
\usepackage{amsthm}
\usepackage{booktabs}
\usepackage{algorithm}
\usepackage{algorithmic}
\usepackage[inkscapelatex=false]{svg}
\usepackage{multirow}
\usepackage{array}
\usepackage{tcolorbox}
\usepackage{pgfmath}
\usepackage{xfp} 
\usepackage{tabularx}
\usepackage{wrapfig}
\usepackage{mdframed}
\usepackage[normalem]{ulem}
\usepackage{makecell}
\usepackage{amssymb}
\usepackage{longtable}
\usepackage{cuted}
\usepackage{listings}
\usepackage{paralist}

\lstdefinelanguage{prolog}{
  keywords={not},
  keywordstyle=\color{blue}\bfseries,
  comment=[l]{\%},
  commentstyle=\color{green!60!black},
  stringstyle=\color{red},
  basicstyle=\ttfamily\footnotesize,
  breaklines=true,
  breakatwhitespace=true,
  showstringspaces=false,
  frame=single,
  backgroundcolor=\color{gray!10},
  numberstyle=\tiny\color{gray},
  rulecolor=\color{gray!30}
}

\soulregister{\ref}{1}
\definecolor{mygreen}{HTML}{E0FEE0}
\definecolor{myred}{HTML}{FEE0E0}

\newcommand{\citeasnoun}[1]{\citeauthor{#1}~\shortcite{#1}}

\title{
  Can LLMs Solve ASP Problems? Insights from a Benchmarking Study 
  
  (Extended Version)
  }

\author{%
Lin Ren$^{1,2}$\and
Guohui Xiao$^{1,2}$\thanks{Corresponding author}\and
Guilin Qi$^{1,2}$\and
Yishuai Geng$^{1,2}$\and
Haohan Xue$^{1,2}$ \\
\affiliations
$^1$School of Computer Science and Engineering, Southeast University, Nanjing, China\\
$^2$Key Laboratory of New Generation Artificial Intelligence Technology and Its Interdisciplinary Applications (Southeast University), Ministry of Education, China \\
\emails
\{renlin, guohui.xiao, gqi, ysgeng, thex1ay\}@seu.edu.cn
}
\begin{document} 

\maketitle

\begin{abstract}
  Answer Set Programming (ASP) is a powerful paradigm for non-monotonic reasoning.
  Recently, large language models (LLMs) have demonstrated promising capabilities in logical reasoning.
  Despite this potential, current evaluations of LLM capabilities in ASP are often limited.
  Existing works normally employ overly simplified ASP programs, do not support negation, disjunction, or multiple answer sets.
  Furthermore, there is a lack of benchmarks that introduce tasks specifically designed for ASP solving.
  To bridge this gap, we introduce ASPBench, a comprehensive ASP benchmark, including three ASP specific tasks: ASP entailment, answer set verification, and answer set computation. 
  Our extensive evaluations on ASPBench reveal that while 14 state-of-the-art LLMs, including \emph{deepseek-r1}, \emph{o4-mini}, and \emph{gemini-2.5-flash-thinking}, perform relatively well on the first two simpler tasks, they struggle with answer set computation, which is the core of ASP solving. 
  These findings offer insights into the current limitations of LLMs in ASP solving. 
  This highlights the need for new approaches that integrate symbolic reasoning capabilities more effectively.
  The code and dataset are available at \url{https://github.com/HomuraT/ASPBench}.
\end{abstract}

\section{Introduction}
\label{sec:introduction}
Answer Set Programming (ASP)~\cite{gelf-lifs-91,niemela1999logic} is a declarative programming paradigm designed for complex knowledge representation and problem-solving tasks. 
A key strength of ASP is its ability to model situations where conclusions must be revised as new information becomes available~\cite{ginsberg1980readings,reiter1988nonmonotonic}, enabling an adaptive and context-sensitive inference process. 
ASP provides a robust computational framework for various forms of logical reasoning, including default reasoning~\cite{defaultReasoning_1980}, and connects with abductive inference~\cite{abductiveInference_1996} and belief revision~\cite{beliefRevision_1997}.

Recently, large language models (LLMs) have demonstrated remarkable capabilities in diverse areas, such as information retrieval~\cite{zhu2023llm4ir}, question answering~\cite{yue2025llm4qa}, and code generation~\cite{jiang2024llm4code}.
Furthermore, their potential in logical reasoning tasks is an active and growing area of research~\cite{liu2025llm4logic}.
The question of whether LLMs possess logical reasoning ability, and if so, to what extent, has been widely explored~\cite{huang2023towards,wang2024can}, particularly in formalisms such as Description Logics (e.g., DL-Lite)~\cite{llm4DL}, Propositional Logic~\cite{chanrulebreakers,chen2024can}, and First-Order Logic~\cite{chen2024can,wang2024can}.
Beyond these monotonic formalisms, ASP offers a powerful framework for non-monotonic reasoning (NMR).
Consequently, recent works increasingly leverage ASP to evaluate the NMR abilities of LLMs.
\citeasnoun{LogicNMR} created a dataset called LogicNMR for evaluating default logic rules, which effectively only covers a simple ASP fragment.
Similarly, \citeasnoun{parmar2024logicbench} introduced LogicBench for evaluating the logical reasoning capabilities of LLMs, notably featuring patterns from non-monotonic logic (such as default reasoning). 
These works focus on evaluating ability of LLMs to perform symbolic reasoning.
In contrast, \citeasnoun{DefeasibleNLI} and \citeasnoun{PenguinsDonotFly} explore reasoning with defeasible information from the implicit background knowledge of LLMs.

\begin{table}[t]
    \resizebox{\columnwidth}{!}{%
    \footnotesize
    \setlength{\tabcolsep}{3pt}%
    \renewcommand{\arraystretch}{1}%
    \begin{tabular}{lccccccc}
    \toprule
    Dataset & Arity & Repr. & \makecell{Ops.} & MAS & ASE & ASV & ASC \\ \midrule
    $\delta $-NLI & N/A & Tex & DN & N/A & \cellcolor{myred}$\times$ & \cellcolor{myred}$\times$ & \cellcolor{myred}$\times$ \\ \midrule
    ProofWriter & 1 & Tex & SN & N/A & \cellcolor{mygreen}$\checkmark$ & \cellcolor{myred}$\times$ & \cellcolor{myred}$\times$ \\ \midrule
    ruletaker & N/A & Tex & SN & N/A & \cellcolor{mygreen}$\checkmark$ & \cellcolor{myred}$\times$ & \cellcolor{myred}$\times$ \\ \midrule
    LogicNMR & 1 & Tex & SN, DN & \cellcolor{myred}Single & \cellcolor{mygreen}$\checkmark$ & \cellcolor{myred}$\times$ & \cellcolor{myred}$\times$ \\ \midrule
    generics-exemplars & N/A & Tex & N/A & N/A & \cellcolor{myred}$\times$ & \cellcolor{myred}$\times$ & \cellcolor{myred}$\times$ \\ \midrule
    LogicBench & 1 & Tex & SN, DN & \cellcolor{myred}Single & \cellcolor{mygreen}$\checkmark$ & \cellcolor{myred}$\times$ & \cellcolor{myred}$\times$ \\ \midrule
    \textbf{ASPBench (Ours)} & \textbf{Any} & \textbf{\makecell{Sym \\ \& Tex}} & \textbf{\makecell{SN, DN, \\ Disj, Cons}} & \cellcolor{mygreen}\textbf{Multiple} & \cellcolor{mygreen}\textbf{$\checkmark$} & \cellcolor{mygreen}\textbf{$\checkmark$} & \cellcolor{mygreen}\textbf{$\checkmark$} \\ \bottomrule
    \end{tabular}%
    } %
    \caption{Comparison of ASPBench with other datasets for ASP reasoning. Columns show: Arity (predicate arity); Repr. (Symbolic/Textual representation); Ops. (Supported ASP Operations: SN - Strong Negation, DN - Default Negation, Disj - Disjunction, Cons - Constraints); MAS (Multiple Answer Sets support). Compared datasets are detailed in \S\ref{sec:related_work}.}
    \label{tab:compare_aspbench_with_others}
\end{table}

However, to the best of our knowledge, there are no benchmarks systematically evaluating the capabilities of LLMs for ASP.
Previous studies have overlooked several key factors: 
\begin{inparaenum}[\itshape (1)]
    \item They often employ overly simplified ASP programs, for example, using predominantly unary predicates, focusing mainly on (stratified) default negation, and typically considering only a single answer set scenario. 
    \item The significant impact of predicate semantics (symbolic vs. descriptive names) on LLM reasoning in ASP—a key differentiator from traditional solvers—remains largely unexplored.
    \item Previous benchmarks do not design evaluation tasks specifically for ASP. 
\end{inparaenum}
    
To fill these gaps, we propose a novel benchmark for ASP, referred to as ASPBench.
ASPBench features a novel framework for the automated generation of diverse ASP problems with varied rule styles and logical operations.
The benchmark itself incorporates support for multiple answer sets, a broader range of ASP operators, and introduces three distinct ASP evaluation tasks: ASP entailment (ASE), answer set verification (ASV), and answer set computation (ASC).
In addition to synthetic data, ASPBench includes real-world ASP programs collected from public sources.
The differences between ASPBench and related datasets, in the context of evaluating ASP style reasoning, are shown in Table~\ref{tab:compare_aspbench_with_others}.

Our extensive experiments on ASPBench, evaluating 14 state-of-the-art LLMs, reveal several key insights:
\begin{inparaenum}[\itshape (1)]
    \item \textbf{The characteristics of ASP programs significantly influence LLM performance}.
    LLMs struggle with complex tasks such as ASC on synthetic ASP programs. 
    This challenge increases when dealing with real-world ASP programs, where performance significantly drops across all evaluated tasks.
    Furthermore, LLM reasoning is highly sensitive to program features, including their syntactic properties (e.g., whether they are positive, stratified, or head-cycle-free) and the number of answer sets.
    \item \textbf{The inherent properties of LLMs and their inference strategies are vital for their ability to solve ASP}.
    Larger LLMs generally achieve higher performance while producing shorter outputs.
    Conversely, for LLMs with comparable parameters, the length of chain-of-thought tokens strongly correlates with ASP solving abilities, underscoring the value of increased test-time computation.
\end{inparaenum}

\section{Related Work}
\label{sec:related_work}

This work is situated within the context of research on LLMs for logical reasoning, their application as logic solvers or code executors, and existing benchmarks for ASP solving.
\subsection{Logical Reasoning with LLMs}
Recently, LLMs have shown a powerful ability in various monotonic logical reasoning tasks, such as Multi-Step Reasoning~\cite{saha2023murmur,fu2023specializing} and Commonsense Reasoning~\cite{tian2023harnessing,perak2024incorporating}. However, LLMs also exhibit notable limitations in reasoning tasks. \citeasnoun{reasoning_limitation_1} showed that LLM understanding of fundamental reasoning rules lags significantly behind human capability. \citeasnoun{ishay2023leveraging} demonstrated the potential of LLMs in generating complex ASP programs through few-shot prompting, but most errors require manual correction. \citeasnoun{reasoning_limitation_symbolic} explored the challenges LLMs face in solving math word problems, while \citeasnoun{reasoning_limitation_rational} demonstrated that LLMs perform considerably worse than neural program induction systems in reasoning tasks. \citeasnoun{reasoning_limitation_DL_Lite} illustrated that LLMs struggle with understanding TBox NI transitivity rules. \citeasnoun{parmar2024logicbench} showed that LLMs do not perform well in logic reasoning, even though they are in single inference rule scenarios.

\subsection{LLMs as Logic Solvers or Code Executors}
Recently, code has been recognized as a powerful tool for LLMs to access and leverage external sources~\cite{yang2024lls_is_wizard}. Meanwhile, there has been growing interest in exploring the role of LLMs as logic solvers and code executors. For example, \citeasnoun{feng2023LLM_logic_solvers} utilized LLMs as Prolog logic solvers to address parsing errors in logic programs. Similarly, \citeasnoun{chen2024LLM_logic_solvers} explored how to guide LLMs in simulating logic solvers to execute Propositional Logic or Satisfiability Modulo Theories (SMT) programs, using natural language, Z3Py~\cite{de2008z3}, or SMT-LIB~\cite{barrett2010smt}. Additionally, \citeasnoun{wang2024z3} demonstrated that LLMs can serve as executors when generated Z3 programs fail during execution, and \citeasnoun{lyu2024llm_be_code_executors} explored the feasibility of using LLMs as Python code executors. Our work focuses on leveraging LLMs as ASP solvers.

\subsection{Benchmarks for ASP Solving with LLMs}
To evaluate the ASP solving ability of language models, several benchmarks have been proposed. 
$\delta$-NLI~\cite{DefeasibleNLI} was introduced for non-monotonic inference by assessing belief changes with new information. 
LogicNMR~\cite{LogicNMR} provides a dataset of textual ASP samples. 
generics-exemplars~\cite{PenguinsDonotFly} focused on reasoning about generics and their exceptions. 
For broader logical reasoning, ProofWriter~\cite{Tafjord2021ProofWriter} was developed for generating natural language proofs, and RuleTakers~\cite{clark2021ruletaker} were created for emulating reasoning over textual rules. 
\citeasnoun{AreLLMsClassicalOrNonmonotonicReasoners} also explored belief stability in generics. 
Additionally, LogicBench~\cite{parmar2024logicbench} provides benchmarks for logical reasoning, encompassing some non-monotonic scenarios. 
\citeasnoun{borroto2024towards} introduced NL2ASP, a two-step architecture that demonstrates the potential for automated symbolic ASP program generation from natural language specifications. 
Our work focuses on a comprehensive evaluation of ASP solving ability of LLMs, using more complex synthetic ASP programs and real-world ASP programs.

\section{Preliminary}
\label{sec:preliminary}

\subsection{Answer Set Programming}
\label{subsec:asp}
In this work, we employ the framework of Answer Set Programming (ASP)~\cite{gelfond1988stable,gelf-lifs-91}.
An ASP program is a set of rules of the following form:
\begin{align*}
    \omega_1(\textbf{x}_1) | \ldots | \omega_k(\textbf{x}_k) \leftarrow 
    &\alpha_1(\textbf{x}_1), \ldots, \alpha_m(\textbf{x}_m),  \\
    &\text{not} \ \alpha_{m+1}(\textbf{x}_{m+1}), ... , \text{not} \ \alpha_n(\textbf{x}_n)
    \label{eq:definition_ASP_logic}
\end{align*}
where each $\omega_i(\textbf{x}_i)$ is an atom and each $\alpha_j(\textbf{x}_j)$ is a literal of the form $\texttt{p}(\textbf{x}_j)$ (positive literal) or $\texttt{-}\texttt{p}(\textbf{x}_j)$ (negative literal), and each $\textbf{x}_i$ or $\textbf{x}_j$ consists of variables and constants.
In ASP, 
``\texttt{not}''  and ``\texttt{-}'' are called the default negation (negation-as-failure) and the classical negation (strong negation), respectively. An ASP program or a rule is ground if there are no variables.  A fact is a ground rule with $n=0$. We often write an ASP problem as a pair $(W,D)$ with $W$ a set of facts, and $D$ a set of rules.

The semantics of ASP are characterized by the notion of answer sets, also known as stable models~\cite{gelfond1988stable}.
The semantics of an ASP program $P=(W,D)$ is defined via the Gelfond-Lifschitz transformation.
Let $S$ be a candidate set of ground atoms.
The reduct $P^S$ of the program $P$ with respect to $S$ is obtained from the set of all ground instances of rules in $D \cup W$ by:

\textit{(1)} Deleting each rule $r$ such that its body contains a default literal $\text{not } \alpha$ and $\alpha \in S$.

\textit{(2)} Deleting all default literals $\text{not } \alpha$ from the bodies of the remaining rules.

The resulting program $P^S$ is a positive (default-negation-free) disjunctive logic program.
A set $S$ is an answer set of $P$ if and only if $S$ is a minimal model of $P^S$.
A model $M$ of $P^S$ is minimal if there is no model $M' \subset M$ of $P^S$.
In this context, facts from $W$ are treated as rules with empty bodies.

% Following our running example, $P_0$ has an answer set $W_0 \cup \{ \text{CanFly(\emph{Tweety})} \}$, $P_1$ has an answer set $W_1 \cup \text{Abnormal(\emph{Tweety}})$, and $P_2$ has an answer set $W_2 \cup \{ \text{CanFly(\emph{Tweety})}\}$.

The ASP paradigm has been implemented in several ASP solvers, e.g., DLV~\cite{dlv2} and Clingo~\cite{gekakasc12a}.

\subsection{Syntactic Classes}
\label{subsec:syntactic_class_definitions}
In this work, we are interested in the following syntactic classes of ASP programs based on their properties:

\textbf{Positive Program}: A program is \emph{positive} if it contains \emph{only} positive atoms (i.e., neither strong negation nor default negation occurs).

\textbf{Stratified Program}: A program is \emph{stratified} when its dependency graph has no directed cycle that traverses a negative edge, i.e. default negation is never involved in recursion~\cite{apt1988towards}.  Positive and strong literals may still form cycles.
For examples, \textit{P1} = \texttt{\{-p :- -q. -q :- -p.\}} is stratified, because it has no recursion through default negation;
\textit{P2} = \texttt{\{p :- not q. q :- not p.\}} is \emph{unstratified} , because the cycle between \texttt{p} and \texttt{q} goes through default negation.

\textbf{Head-Cycle-Free~(HCF) Program}: A program is \emph{head-cycle-free} when its dependency graph is acyclic, i.e., it contains no directed cycle consisting solely of positive or negated atoms~\cite{ben1994propositional}.  The presence of multiple atoms in a disjunctive head is allowed; what is prohibited is any positive recursion.
For examples, \textit{P3} = \texttt{\{a | b :- d. c :- a. c :- b. d.\}} is HCF, because no positive cycle exists;
\textit{P4} = \texttt{\{a | b :- c. c :- a. c :- b.\}} is \emph{not} HCF, because the positive dependency graph contains the cycle \texttt{a}\,$\rightarrow$\,\texttt{c}\,$\rightarrow$\,\texttt{a} (and similarly for \texttt{b}), i.e.\ a positive recursion.

\subsection{Task Definitions}
\label{subsec:aspbench_tasks}
We define three distinct downstream tasks to evaluate the reasoning capabilities of models on ASP solving. These tasks cover different aspects of ASP reasoning:

\textit{(1)} \textbf{ASP Entailment (ASE):} Given an ASP program $P$ and a ground atom \texttt{a}. The task is to determine the truth state of \texttt{a} within the answer set $S$. The expected output is one of three possibilities: \texttt{true} (if \texttt{a} $\in$ $S$), \texttt{false} (if $\text{-}\texttt{a}$ $\in$ $S$), or \texttt{unknown} (if neither \texttt{a} $\in$ $S$ nor $\text{-}\texttt{a}$ $\in$ $S$).

\textit{(2)} \textbf{Answer Set Verification (ASV):} Given an ASP program $P$ (guaranteed to have one or more answer sets, potentially generated using disjunction in rule heads) and a candidate set of ground atoms $C$, the task is to determine whether $C$ is an actual answer set of $P$. 

\textit{(3)} \textbf{Answer Set Computation (ASC):} Given an ASP program $P$, the task is to compute and return one of its correct answer sets.

\section{ASPBench}
\label{sec:methodology}

\begin{figure*}[ht!]
  \centering
  \includegraphics[width=1\linewidth]{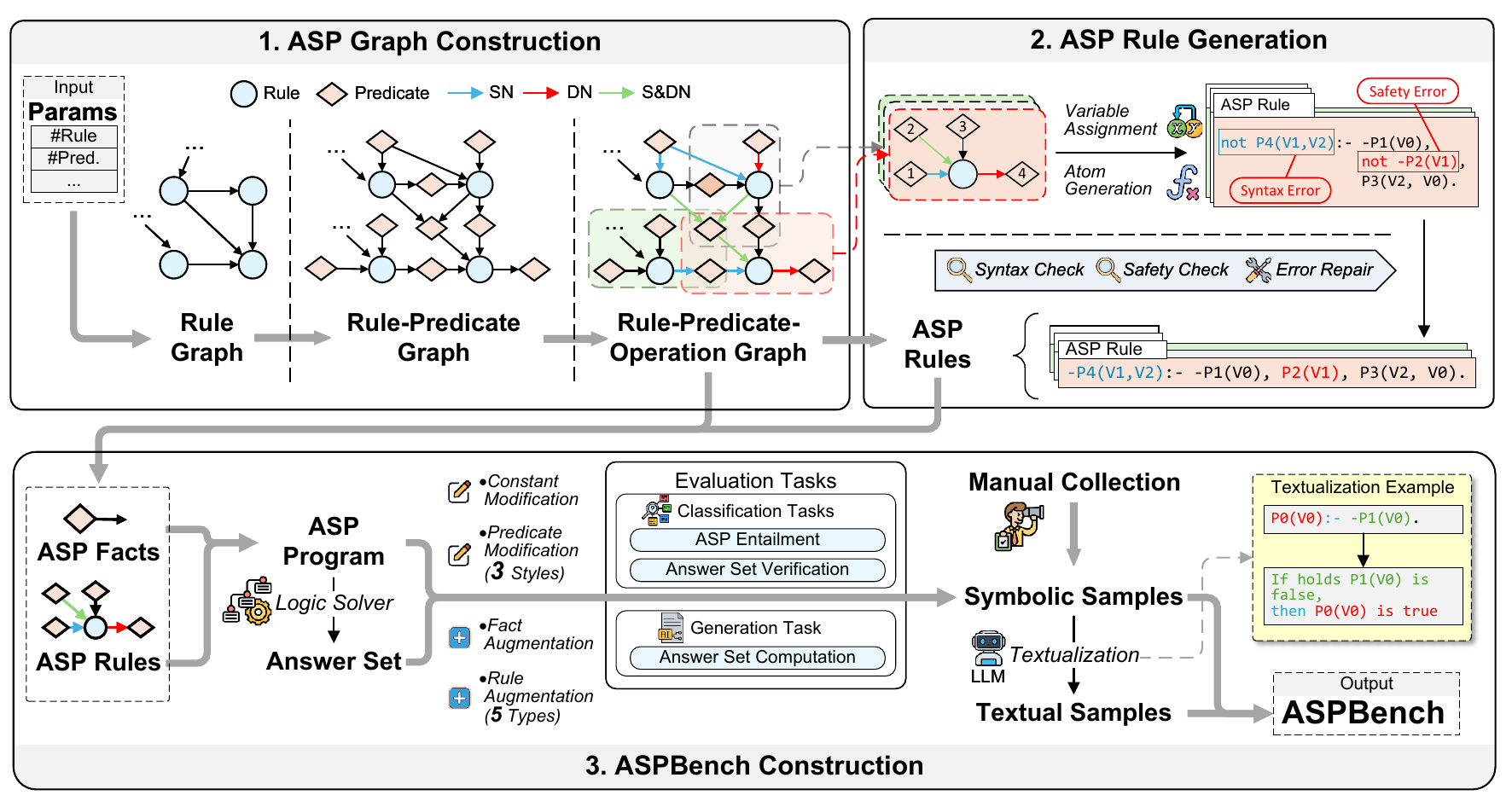}
  \caption{The generation framework for ASPBench}
  \label{fig:overall_flow}
\end{figure*}

We introduce ASPBench through a systematic three-stage generation pipeline (Figure~\ref{fig:overall_flow}), which enables precise control over sample complexity and diversity. 
\begin{inparaenum}[\itshape(1)]
\item \textbf{ASP Graph Construction} (\S~\ref{subsec:asp_graph_definition}, \ref{subsec:asp_graph_construction}): We construct an ASP Graph that represents the logical dependency structure. This graph-based representation allows us to control key properties such as reasoning depth and structural complexity. 
\item \textbf{ASP Rule Generation} (\S~\ref{subsec:asp_rule_generation}): We transform the ASP Graph into concrete, syntactically valid ASP rules.s 
\item \textbf{ASPBench Construction} (\S~\ref{subsec:aspbench_construction}, \ref{subsec:task_design_methodology}): We construct the symbolic benchmark tailored to our three target tasks (i.e., ASP Entailment, ASP Verification, and ASP Computation) and generate the corresponding textual samples. 
\end{inparaenum}

This section mainly introduces the architecture and key steps. The detailed procedures, hyperparameters, and other specific contents are provided in Appendix A.

\subsection{Definition of ASP Graph}
\label{subsec:asp_graph_definition} 
Similar to the Rete algorithm~\cite{forgy1989rete}, we use a graph structure to represent a logical program.
An \textbf{ASP Graph} is a directed graph that serves as a structured and detailed representation of an ASP program. 

The graph utilizes two types of nodes:
\textbf{Rule Nodes ($N_R$)} represent the individual rules within the ASP program.
Each rule node corresponds to a specific rule.
\textbf{Predicate Nodes ($N_P$)} represent the unique predicates used in the program.
Each predicate node corresponds to a distinct predicate.

Moreover, four types of edges define the logical operations between nodes:
\begin{inparaenum}[\itshape (1)]
  \item \textbf{Default (P):} The predicate is not subject to any negation operators.
  \item \textbf{Strong Negation (SN):} The predicate is subject to strong negation.
  \item \textbf{Default Negation (DN):} The predicate is subject to default negation.
  \item \textbf{Combined Negation (SN\&DN):} The predicate is subject to both default and strong negation.
\end{inparaenum}
The term ``ASP Graph'' in this work refers to this Rule-Predicate-Operation Graph. 

\subsection{ASP Graph Construction}
\label{subsec:asp_graph_construction}

The initial step in the framework is ASP Graph Construction, which aims to generate an ASP Graph according to hyperparameters (e.g., number of rules, predicates, arity, etc.). 
This process, illustrated in the top-left panel of Figure~\ref{fig:overall_flow}, unfolds in three stages, progressively adding detail to the graph:

\textbf{Rule Graph:} 
This initial graph consists of a predefined number of rule nodes ($N_R$). 
Directed edges are then randomly generated between these nodes, up to a specified total number of edges. 
This random generation is guided by several constraints: \textit{(a)} the resulting graph must be a directed acyclic graph with a single terminal node (no outgoing edges), and \textit{(b)} every rule node must have at least one connecting edge. 
An edge $N_R^i \rightarrow N_R^j$ in this graph represents a dependency, where the head predicate of rule $N_R^i$ appears in the body of rule $N_R^j$.

\textbf{Rule-Predicate Graph:} 
This stage refines the Rule Graph by introducing predicate nodes ($N_P$) and associated edges. 
First, specific predicates are introduced:
\begin{itemize}
    \item For each edge $N_R^i \rightarrow N_R^j$ in the Rule Graph, a predicate $N_P^k$ is inserted, forming $N_R^i \rightarrow N_P^k \rightarrow N_R^j$.
    \item For each source rule node $N_R^{source}$ (no incoming $N_R \rightarrow N_R$ edges), an input predicate $N_P^{in}$ is added with an edge $N_P^{in} \rightarrow N_R^{source}$.
    \item For the terminal rule node $N_R^{term}$ (no outgoing $N_R \rightarrow N_R$ edges), an output predicate $N_P^{out}$ is added with an edge $N_R^{term} \rightarrow N_P^{out}$.
\end{itemize}
Subsequently, the graph is extended by creating a specified number of new predicate nodes and edges.

\textbf{Rule-Predicate-Operation Graph:} The final stage refines the Rule-Predicate Graph by randomly assigning types to edges based on the probability distribution. 
The types of edges are P, SN, DN, or SN\&DN.
To ensure the derivability of the entire graph structure and facilitate quantitative analysis (e.g., controlling reasoning chain length), we set the incoming and outgoing edges to a unified type for each predicate node during generation. 

Furthermore, to ensure structural diversity and avoid redundancy in the dataset, we keep only one sample from those with similar graph structures. 

\subsection{ASP Rule Generation}
\label{subsec:asp_rule_generation}

Following the construction of the Rule-Predicate-Operation Graph, the ASP Rule Generation step translates the graph into concrete ASP rules. As depicted in the top-right panel of Figure~\ref{fig:overall_flow}, this process involves several stages for each rule node in the graph:

\textit{(1)} \textbf{Input Collection:} For a given rule node in the graph, all adjacent predicate nodes and the corresponding edge information (both type and direction) are gathered. This information determines which predicates constitute the head and body of the rule and the logical operations involved.

\textit{(2)} \textbf{Initial Rule Formulation:} Based on the collected information, two sub-steps occur:

\emph{Variable Assignment:}
This step assigns appropriate variables to the atoms, determining the arity (n-ary) of each predicate. 
The initial arity (n-ary) for each predicate is randomly assigned within a predefined range. 
Once assigned, the arity of a predicate is maintained consistently whenever it appears in the program. 
If a predicate has been assigned variables previously, its arity is maintained. 
Additionally, to ensure the rationality of the generated rules, we ensure that the variables of one predicate satisfy the following two conditions:
\textit{(a)} At least one of the variables must occur in the other predicates (either the body or the head).
\textit{(b)} All head predicate variables must be present in the body.

\emph{Atom Generation:}
This step converts predicate nodes into ASP atoms along with the assigned variables, using P-style identifiers (e.g., ``\texttt{P1(V0)}'').

These atoms are then assembled into an ASP rule based on the edge directions and types. 
If a rule node in the graph has multiple outgoing edges representing potential head predicates, the generation process can be adjusted according to the requirements of the task.
By default, when generating programs intended for the ASE task, which requires a unique answer set, disjunctions are avoided and multiple rules with the same body are produced (e.g., ``\texttt{a :- b.}'', ``\texttt{c :- b.}''). 
Conversely, for ASV and ASC tasks involving multiple answer sets, rules with disjunctions (e.g., ``\texttt{a | c :- b.}'') will be generated. Figure~\ref{fig:overall_flow} shows an example of an ASP rule.

\textit{(3)} \textbf{Error Checking and Repair:} 
The initial ASP rules may contain errors. Therefore, we carry out the following checks and repairs:

\emph{Syntax Check:}
Ensuring that the rule adheres to the syntax regulations of ASP (e.g., ``\texttt{not}'' cannot appear in the rule head).

\emph{Safety Check:} 
Verifying that the rule is safe, meaning every variable appearing in the head or in a default negated literal also appears in a body literal that is not default negated (i.e., a positive or strongly negated literal).

\emph{Error Repair:} 
The ASP rules are checked for syntax and safety errors using DLV2.
If any errors are detected, an automated repair process will be carried out.
This process attempts to make the rule valid by making minimal modifications, by applying a ``negation flip'' to predicate atoms. This operation simultaneously toggles both the strong negation status and the default negation status of an atom. For instance, applying this flip to ``\texttt{not p}'' changes it to ``\texttt{-p}'' ; similarly, applying it to ``\texttt{not -p}'' changes it to ``\texttt{p}''.
The repair process iteratively modifies predicate negations, attempting progressively more changes until the rule passes the check.
If a valid rule cannot be obtained after a set number of attempts, the sample generation for this specific rule will be discarded.

This step produces a set of syntactically correct and safe ASP rules, derived from the graph representation and ready for the ASPBench Construction phase.

\subsection{ASPBench Construction}
\label{subsec:aspbench_construction}

The final step, ASPBench Construction, synthesizes the symbolic and textual ASP samples, as illustrated in the bottom panel of Figure~\ref{fig:overall_flow}. This process involves four stages:

\textit{(1)} \textbf{ASP Program Formulation:} Initial ASP facts are generated from input predicate nodes (those lacking incoming edges). 
The truth value (``\texttt{p.}'' or ``\texttt{-p.}'') of a fact is determined by the unified type of its predicate node: types \textbf{P} and \textbf{SN} directly yield ``\texttt{p.}'' and ``\texttt{-p.}'' respectively. 
If the type includes default negation (\textbf{DN} or \textbf{SN\&DN}), a ``negation flip'' operation is applied to determine the final truth value.

\textit{(2)} \textbf{Diversification and Augmentation:} To enhance dataset diversity, several modifications are applied: % : maybe change to Fact and Rule Augmentation

\emph{Fact and Rule Augmentation:} 
To further enhance the diversity and complexity, we introduce 5 distinct types of additional rules to the ASP program.
These rules are generated based on the target predicates (predicate nodes without outgoing edges) $P_t$, other predicates (predicate nodes with outgoing edges) $P_o$, or newly introduced predicates $P_n$.
The structures of the additional rules are as follows:
\begin{inparaenum}[\itshape (a)]
  \item The rule head contains predicates from $P_t$.
  \item The rule body contains predicates from $P_t$.
  \item Both the head and body consist solely of predicates from $P_o$.
  \item Part of the predicates from $P_n$ are in the head and/or body, others are from $P_o$.
  \item Both the head and body consist solely of predicates from $P_n$.
\end{inparaenum}
Furthermore, if $P_n$ is introduced, additional facts regarding it will be added when generating rules.

\emph{Constant Modification:} 
Constants in facts will be replaced with randomly generated names to vary the grounding (e.g., ``\texttt{P1(V1)}'' $\rightarrow$ ``\texttt{P1("Tweety")}'').

\emph{Predicate Styling:} 
Predicate descriptions are modified using three distinct styles: 
\begin{inparaenum}[\itshape (a)]
  \item simple P-style identifiers (P1, P2,...); 
  \item related concepts drawn from ConceptNet triples. For example, replacing predicates ``\texttt{a(V0)}'' and ``\texttt{b(V0)}'' in ``\texttt{a(V0) :- -b(V0).}'' with ``\texttt{flying}'' and ``\texttt{bird}'' based on the ConceptNet relation ``\texttt{bird, CapableOf, flying}'', even if the resulting rule lacks real-world logical validity, such as ``\texttt{flying(V0) :- -bird(V0).}''; or 
  \item random concepts are selected from the ConceptNet and used as predicate descriptions.
\end{inparaenum}

\textit{(3)} \textbf{Symbolic Sample Generation:} 
The modified ASP program will be executed using DLV2 and then saved if answer sets are produced successfully.
These answer sets will be parsed and, together with the program itself, will form the symbolic ASP sample.

\textit{(4)} \textbf{Textualization:} The symbolic ASP sample (facts, rules, and potentially a query) is converted into a natural language description using template-based conversion rules, which is then proofread by an LLM\footnote{We use gpt-4o-mini during textualization.} to fix grammatical and punctuation errors\footnote{We conducted a human check to ensure the quality of textual samples.}. 
To mostly keep the logical structure during textualization, we use a simple but precise style. 
For instance, positive facts like ``\texttt{p(X).}'' are described as ``p(X) is true'', while negative facts like ``\texttt{-p(X).}'' are rendered as ``p(X) is explicitly false''. 
Rules such as ``\texttt{h :- b1, not b2.}'' are converted into conditional statements like ``If b1 is true and there is no evidence that b2 is true, then h is true''. 
Details about the textualization and the prompt are provided in the supplementary material (Appendix B, C).

This step produces a pair of corresponding symbolic and textual ASP samples, which constitute the samples of the ASPBench dataset.

\subsection{Task Design}
\label{subsec:task_design_methodology}
ASPBench is designed to evaluate LLMs on three core ASP reasoning tasks: ASP entailment, answer set verification, and answer set computation.
This section provides details of the settings and generation process used by ASPBench to create samples for each task.

\subsubsection{ASP Entailment}
For each ASE sample, we ensure that the ASP program $P$ does not contain any disjunctions and has only one unique answer set $S$. 
The query atom $a$ is positive and is generated from the terminal predicate node $P_t$. 
The label is the truth value of $a$ with respect to $S$.

\subsubsection{Answer Set Verification}
For each ASV sample, the program $P$ is allowed to include disjunctions and is guaranteed to have at least one answer set.
The label for the candidate set $C$ is randomly assigned as either \texttt{true} or \texttt{false}.
When the label is \texttt{true}, $C$ is randomly selected from one of the answer sets of $P$.
When the label is \texttt{false}, $C$ is generated by taking a correct answer set of $P$ and applying exactly one of the following modifications:
\begin{itemize}
    \item \emph{Flip Negation:} A fact is randomly selected from a correct answer set, and its negation status is flipped (e.g., ``\texttt{p.}'' becomes ``\texttt{-p.}'', or ``\texttt{-p.}'' becomes ``\texttt{p.}'').
    \item \emph{Delete Fact:} A single fact is randomly removed from a correct answer set.
    \item \emph{Add Modified Fact (Constants):} A fact is randomly selected from a correct answer set, its constants are altered, and this newly formed fact is added to the set.
    \item \emph{Add Modified Fact (Predicate):} A fact is randomly selected from a correct answer set, its predicate name is changed, and this newly formed fact is added to the set.
\end{itemize}

\subsubsection{Answer Set Computation}
For each ASC sample, the program $P$ is allowed to include disjunctions and is guaranteed to have at least one answer set.
Moreover, all the answer sets of $P$ are saved.
The output of the model for the ASC task is then evaluated for correctness by checking whether it matches one of these answer sets.

In summary, the ASE task in ASPBench is tailored for scenarios with unique answer sets by constraining rule generation.
Conversely, the ASV and ASC tasks embrace the possibility of multiple answer sets, often involving head disjunction, thereby providing a comprehensive evaluation of reasoning over more complex ASP features.
\section{Experiment}
\label{sec:experiment}
\label{sec:exp}
\subsection{Dataset Statistics}
\label{subsec:dataset_statistics}
ASPBench benchmark includes three datasets for evaluating the ASP solving capabilities of LLMs:
ASP Entailment (ASE), Answer Set Verification (ASV), and Answer Set Computation (ASC).
Each dataset consists of 1,000 samples, pairing symbolic ASP programs with textual descriptions.
Table~\ref{tab:dataset_statistics} details their statistics.
Note that constraints are often ineffective in synthetic programs, so ASPBench only incorporates them in manually collected ASP programs.

Moreover, in addition to synthetic ASP programs, we collected 47 symbolically represented classic ASP programs from various categories, as shown in Table~\ref{tab:classic_problems_statistics}. 
To enhance diversity and prevent LLMs from providing correct answers by memorising programs, we extend each program by: 
\begin{inparaenum}[\itshape (1)]
    \item scaling up problem instances;
    \item converting problems to P-style format.
\end{inparaenum}

\begin{table}[ht]
\centering
\resizebox{\columnwidth}{!}{%
\setlength{\tabcolsep}{15pt}
\renewcommand{\arraystretch}{0.7}
\begin{tabular}{lccc}
\hline
\textbf{Statistic} & \textbf{ASE} & \textbf{ASV} & \textbf{ASC} \\
\hline
Program Size & & & \\
Avg. Rules & 10.72 & 15.71 & 15.47 \\
Avg. Facts & 9.77 & 14.36 & 14.23 \\
Pred. Arity & 0-3 & 0-3 & 0-3 \\
Max. Chain & 1-7 & - & - \\
\hline
Syntactic Classes & & & \\
Positive & 10.1\% & 9.2\% & 10.3\% \\
Stratified & 34.5\% & 34.8\% & 35.7\% \\
Head-Cycle-Free & 42.9\% & 44.3\% & 44.9\% \\
\hline
Answer Sets & & & \\
Avg. Count & 1.00 & 3.33 & 3.44 \\
Avg. Facts/Set & - & 22.70 & 21.94 \\
\hline
Label Dist. & & & \\
True & 41.3\% & 50.7\% & - \\
False & 32.1\% & 49.3\% & - \\
Unknown & 26.6\% & - & - \\
\hline
Pred. Style & & & \\
P-style & 32.7\% & 32.7\% & 32.5\% \\
Related & 33.8\% & 32.5\% & 32.4\% \\
Random & 33.5\% & 34.8\% & 35.1\% \\
\hline
\end{tabular}
}
\caption{Statistics of the ASPBench dataset across different tasks (ASE, ASV, ASC).}
\label{tab:dataset_statistics}
\end{table}

\begin{table*}[ht]
\centering
%\resizebox{\columnwidth}{!}
\small
\resizebox{\textwidth}{!}{%
\setlength{\tabcolsep}{8pt}
\renewcommand{\arraystretch}{1}
\begin{tabular}{lcl}
\hline
\textbf{Category} & \textbf{Number} & \textbf{Characteristics and Representative Examples} \\
\hline
\multirow{2}{*}{\makecell[l]{Basic Logic Reasoning (BLR)}} & \multirow{2}{*}{\makecell{8}} & Classic logic puzzles requiring deductive reasoning and constraint satisfaction. \\
& & \textit{Examples:} Zebra puzzle, Who killed Agatha, Safe cracking puzzle, etc. \\
\hline
\multirow{2}{*}{\makecell[l]{Combinatorial Search \\ \& Optimization (CSO)}} & \multirow{2}{*}{\makecell{15}} & Tasks involving state-space exploration, pattern matching, and combinatorial optimization. \\
& & \textit{Examples:} N-Queens, Sudoku, Magic square, Minesweeper, etc. \\
\hline
\multirow{2}{*}{\makecell[l]{Constraint  Satisfaction \\ \& Scheduling (CSS)}} & \multirow{2}{*}{\makecell{8}} & Resource allocation and temporal scheduling problems under constraints. \\
& & \textit{Examples:} Map coloring, Job scheduling, Traffic light coordination, etc. \\
\hline
\multirow{2}{*}{\makecell[l]{Mathematical \& Set \\ Problems (MSP)}} & \multirow{2}{*}{\makecell{16}} & Problems involving numerical computations and set-based reasoning of varying complexity. \\
& & \textit{Examples:} Set covering, Euler Problem, Subset sum, Prime numbers, etc. \\
\hline
\end{tabular}
}
\caption{Overview of manually collected ASP problems grouped by problem-solving paradigm and reasoning characteristics.}
\label{tab:classic_problems_statistics}
\end{table*}

\subsection{Evaluation setup}
\subsubsection{Models}
To evaluate the reasoning capability of LLMs using the ASPBench dataset, we conducted experiments on 14 LLMs. These models are categorized as follows:
\begin{inparaenum}[\itshape (1)]
    \item \textit{General LLMs}: \emph{qwen2.5-7b}, \emph{qwen2.5-14b}~\cite{yang2025qwen25}, \emph{glm-4-flash}~\cite{glm2024chatglm}, \emph{gpt4o-mini}~\cite{gpt4o-mini}, \emph{gpt-4o}~\cite{gpt4o}, \emph{claude-3-haiku}~\cite{claude-3-haiku}, \emph{deepseek-v3}~\cite{liu2024deepseekv3}, and \emph{gemini-2.5-flash-nothinking}~\cite{Doshi2025GeminiFlash}.
    \item \textit{Reasoning-Optimized LLMs}: \emph{qwen3-8b}, \emph{qwen3-14b}~\cite{yang2025qwen3}, \emph{o3-mini}~\cite{openai2025o3mini}, \emph{o4-mini}~\cite{openai2025o4mini}, \emph{deepseek-r1}~\cite{guo2025deepseekr1}, and \emph{gemini-2.5-flash-thinking}~\cite{Doshi2025GeminiFlash}.
\end{inparaenum}
For each task, we use the same prompt across all LLMs. 
The detailed prompts used in experiments are shown in Appendix~C.

\begin{table*}[htbp]
    \centering
    \resizebox{\textwidth}{!}{
    \setlength{\tabcolsep}{6pt}
    \renewcommand{\arraystretch}{0.5}
    \begin{tabular}{l|cc|ccc|cc|ccc|c|cc|ccc|c}
      \toprule
      \multirow{2}{*}{Model} & \multicolumn{5}{c}{ASP Entailment (F1)} & \multicolumn{6}{c}{Answer Set Verification (F1)} & \multicolumn{6}{c}{Answer Set Computation (EM)} \\
      \cmidrule(lr){2-6} \cmidrule(lr){7-12} \cmidrule(lr){13-18}
      & Sym & Tex & P-style & RanW & RelW & Sym & Tex & P-style & RanW & RelW & RealP & Sym & Tex & P-style & RanW & RelW & RealP \\
      \midrule
      \multicolumn{18}{l}{\textit{General LLMs:}} \\
      gpt-4o-mini   & 0.347 & 0.386 & 0.373 & 0.372 & 0.348 & 0.410 & 0.473 & 0.456 & 0.434 & 0.432 & 0.623 & 0.021 & 0.028 & 0.025 & 0.027 & 0.022 & 0.007 \\
      glm-4-flash   & 0.322 & 0.340 & 0.318 & 0.360 & 0.312 & 0.505 & 0.509 & 0.499 & 0.512 & 0.505 & 0.461 & 0.009 & 0.019 & 0.011 & 0.021 & 0.009 & 0.021 \\
      gpt-4o        & 0.563 & 0.651 & 0.577 & 0.631 & 0.606 & 0.519 & 0.549 & 0.550 & 0.515 & 0.533 & 0.518 & 0.083 & 0.083 & 0.069 & 0.104 & 0.074 & 0.028 \\
      claude-3.5-haiku            & 0.380 & 0.468 & 0.384 & 0.431 & 0.453 & 0.499 & 0.536 & 0.494 & 0.518 & 0.537 & 0.449 & 0.090 & 0.145 & 0.102 & 0.145 & 0.102 & 0.014 \\
      qwen2.5-7b    & 0.328 & 0.283 & 0.268 & 0.337 & 0.309 & 0.534 & 0.518 & 0.496 & 0.549 & 0.522 & 0.540 & 0.014 & 0.016 & 0.015 & 0.026 & 0.005 & 0.014 \\
      \rowcolor{blue!10} qwen2.5-14b   & 0.471 & 0.526 & 0.434 & 0.522 & 0.535 & 0.552 & 0.548 & 0.542 & 0.515 & 0.592 & 0.389 & 0.063 & 0.082 & 0.080 & 0.081 & 0.057 & 0.014 \\
      \rowcolor{green!10} deepseek-v3   & 0.674 & 0.708 & 0.621 & 0.733 & 0.716 & 0.547 & 0.566 & 0.549 & 0.561 & 0.553 & 0.547 & 0.256 & 0.158 & 0.222 & 0.218 & 0.181 & 0.064 \\
      \rowcolor{yellow!20} gemini-2.5-flash-nothinking & 0.890 & 0.832 & 0.849 & 0.861 & 0.871 & 0.783 & 0.786 & 0.792 & 0.791 & 0.768 & 0.621 & 0.065 & 0.109 & 0.071 & 0.091 & 0.099 & 0.074 \\
      \midrule
      \multicolumn{18}{l}{\textit{Reasoning-Optimized LLMs:}} \\
      qwen3-8b      & 0.682 & 0.911 & 0.762 & 0.811 & 0.813 & 0.659 & 0.666 & 0.669 & 0.629 & 0.690 & 0.596 & 0.167 & 0.308 & 0.240 & 0.244 & 0.228 & 0.057 \\
      \rowcolor{blue!10} qwen3-14b     & 0.850 & 0.958 & 0.886 & 0.904 & 0.920 & 0.692 & 0.714 & 0.707 & 0.698 & 0.702 & 0.666 & 0.290 & 0.517 & 0.406 & 0.413 & 0.397 & 0.092 \\
      \rowcolor{green!10} deepseek-r1   & 0.976 & 0.977 & 0.978 & 0.967 & 0.984 & 0.873 & 0.809 & 0.838 & 0.836 & 0.849 & 0.674 & 0.817 & 0.297 & 0.551 & 0.581 & 0.537 & 0.149 \\
      \rowcolor{yellow!20} gemini-2.5-flash-thinking   & 0.967 & 0.870 & 0.909 & 0.923 & 0.922 & 0.794 & 0.775 & 0.775 & 0.778 & 0.796 & 0.521 & 0.244 & 0.234 & 0.206 & 0.258 & 0.257 & 0.171 \\
      o3-mini       & 0.982 & 0.984 & 0.986 & 0.980 & 0.984 & 0.859 & 0.818 & 0.840 & 0.831 & 0.843 & 0.759 & 0.600 & 0.531 & 0.531 & 0.590 & 0.574 & 0.264 \\
      o4-mini       & 0.972 & 0.965 & 0.964 & 0.968 & 0.973 & 0.800 & 0.823 & 0.809 & 0.812 & 0.812 & 0.702 & 0.645 & 0.604 & 0.591 & 0.656 & 0.624 & 0.239 \\
      \midrule
      \textit{Avg.} & 0.672 & 0.704 & 0.665 & 0.700 & 0.696 & 0.645 & 0.649 & 0.644 & 0.641 & 0.652 & 0.576 & 0.240 & 0.224 & 0.223 & 0.247 & 0.226 & 0.086 \\
      \bottomrule
    \end{tabular}
    }
    \caption{\label{tab:overall_performance} Overall performance of different LLMs on various reasoning tasks using the ASPBench datasets and real-world ASP benchmarks.
    The table details F1 scores and EM across different input styles.
    Abbreviations: Sym (Symbolic representation), Tex (Textual representation), P-style (Program-style representation), RanW (Random concepts from ConceptNet), RelW (Related concepts from ConceptNet triples), RealP (Real-world ASP Programs).}

\end{table*} 

\subsubsection{Metrics}

We evaluate performance using task-specific metrics: for ASP entailment and answer set verification, we use the macro-F1 score, which treats all classes equally to mitigate biases from imbalanced label distributions; for answer set computation, we employ Exact Match (EM), defining a sample as correct if any predicted answer set exactly matches any of the ground truth answer sets.

\subsubsection{Implementation details}
Inspired by \citeasnoun{zheng2023judging} and \citeasnoun{tam2024speakfreely}, we do not forcefully restrict the output format of LLMs during reasoning to minimize potential interference with their actual reasoning capabilities.
Instead, we employ \emph{gpt-4o-mini} to convert the raw outputs into structured JSON format for automated evaluation.
For the Answer Set Computation task, we conduct an alignment process before evaluation.
This process maps predicted facts to ground truth answer sets, handling model outputs that may not strictly follow standard answer set representations.
Moreover, we use the latest version of DLV,  \emph{DLV2}\footnote{\url{https://dlv.demacs.unical.it/}}, to validate the correctness of the symbolic samples in ASPBench. 
Details about the prompts for each task are provided in the supplementary material (Appendix C).

\subsection{Main Results}
We report the main findings through the following four questions:

\subsubsection{(1) How LLMs perform on ASP solving in general?} 
Our analysis of Table~\ref{tab:overall_performance} demonstrates that current LLMs have significant limitations when it comes to solving ASP. 
Their performance is generally poor and depends heavily on task structure and input modality.

From the results, we can observe that a \textbf{steep performance cliff exists with task complexity}.
While LLMs achieve moderate F1 scores in ASE (68.8\% for the Sym/Tex average) and ASV (64.7\% for the Sym/Tex average), further analysis reveals significant differences in performance.
Specifically, ASV is a binary classification task and therefore has a higher random baseline (e.g., 0.5 F1 score) than the three-class ASE.
Furthermore, the peak performance achieved in ASE (e.g., 98.2\% by o3-mini on Sym input) significantly outperforms that of ASV (e.g., 87.2\% by deepseek-r1 on Sym input).
These factors suggest that the performance gap between LLMs on these tasks is larger than the average scores suggest, and ASV is therefore a more challenging task.
Moreover, their performance \textit{drops sharply} in the more complex and practically vital \textbf{ASC} task.
The average EM score for ASC is 23.2\% in the synthetic dataset and drops further to 8.6\% in real-world ASP programs.
This highlights the significant challenges involved in generating complete and precise multi-step logical reasoning.

For different types of LLMs, \textbf{Reasoning-Optimized LLMs achieve significant improvements}. 
Compared to \emph{deepseek-v3} which achieves an EM of 20.7\% in ASC, \emph{deepseek-r1} achieves an EM of 55.7\%, a substantial improvement.  
Similar improvements are observed in real-world ASP programs, where \emph{qwen3-14b} shows a notable 9.2\% EM compared to just 1.4\% for \emph{qwen2.5-14b}.
However, even these improved models still present a gap for reliable practical application. 
This highlights that, despite beneficial optimization strategies, the capability of current LLMs to robustly solve these critical and complex reasoning tasks \textbf{requires further significant improvement}.

\subsubsection{(2) How symbolic and textual representations influence the performance of LLMs in ASP solving?}
As shown in Table~\ref{tab:overall_performance} (Avg. row), our results indicate that the relative performance of LLMs on symbolic versus textual representations is \textbf{highly task-dependent}. 

LLMs generally perform better with textual input than symbolic input for classification tasks such as ASE (Tex: 70.4\% vs Sym: 67.2\% F1) and ASV (Tex: 64.9\% vs Sym: 64.5\% F1).
This suggests that their reasoning is better aligned with the linguistic patterns of these contexts.
Conversely, for the \textbf{ASC task}, symbolic inputs show a slightly higher average EM (Sym: 24.0\% vs Tex: 22.4\%). 
This difference may partly arise from the \textbf{challenge of aligning the natural language descriptions of textual outputs with structured ground truths}, which is an issue that is less prevalent for symbolic formats.

\subsubsection{(3) How do the naming styles of predicates influence the ASP solving ability of LLMs?}
The choice of predicate representation style \textbf{markedly influences LLM reasoning}. 

Across all tasks, average results in Table~\ref{tab:overall_performance} indicate that \textbf{LLMs reason more effectively with lexicalized predicates (RanW, RelW) than with simple P-style identifiers}. 
Specifically, P-style representations are consistently outperformed by at least one of the lexicalized predicate styles in each task category. 
For instance, in ASP entailment, the F1 score of P-style (66.5\%) is surpassed by RanW (70.0\%) and RelW (69.6\%).
This implies that predicates with some semantic grounding, whether they are randomly selected concepts from ConceptNet (as in RanW) or related concepts from ConceptNet triples (as in RelW), provide \textbf{more accessible anchors for reasoning} compared to simple identifiers (e.g., ``P1'', ``P2'' for P-style). 
This highlights a \textbf{sensitivity to how predicates are represented}, beyond just the choice between symbolic and textual inputs.

\subsubsection{(4) How do LLMs perform in solving real-world ASP problems, compared to synthetic samples?}
Real-world ASP programs have similar or even fewer rules to synthetic samples, but they have more complex logical structures, such as constraints and iterative rules.

\begin{figure}[h!]
    \centering
    \includegraphics[width=0.6\linewidth]{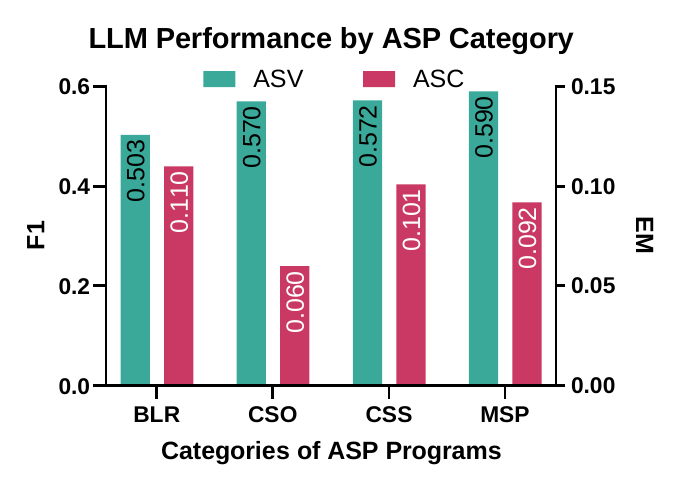}
    \caption{\label{fig:classic_asp_category_performance}LLM performance on four real-world ASP problem categories.}
\end{figure}

Compared with synthetic samples, the performance of LLMs on real-world ASP programs reveals a \textbf{notable performance gap}, with F1 scores dropping from 64.7\% to 57.6\% in ASV and EM scores plummeting from 23.2\% to 8.6\% in ASC (see the ``RealP'' columns in Table~\ref{tab:overall_performance}).
For different categories, as shown in Figure~\ref{fig:classic_asp_category_performance}, the two worst performing categories are Basic Logic Reasoning (BLR) and Combinatorial Search \& Optimization (CSO), with the lowest F1 and EM scores, respectively.
This denotes that current LLMs are almost unable to solve complex ASP problems.

Overall, these findings suggest that, despite their potential for solving basic ASP problems, \textbf{current LLMs lack the robust logical reasoning capabilities required for complex practical applications.}

\subsection{Fine-grained Analysis}
To investigate the limitations of LLMs in ASP solving, we perform fine-grained analysis on ASPBench. Case studies for each task are shown in the supplementary material (Appendix F).

\subsubsection{ASP Entailment}
For ASP entailment (Figure~\ref{fig:sankey_ASE_and_ASV} (a)), we observe that direct flips between the \texttt{true} and \texttt{false} states are rare; most errors arise from cases whose label is \texttt{true} or \texttt{false} but are predicted as \texttt{unknown}. 
Moreover, the proportion of \texttt{unknown} climbs from 26.6\% in the ground truth to 47.2\% in predictions. These observations reveal the following insights:
\begin{inparaenum}[\itshape (1)]
    \item \textbf{Risk-averse bias.} LLMs would rather output \texttt{unknown} than risk making a polarity mistake, prioritising a lower risk of blatant contradiction over recall.
    \item \textbf{Stricter ternary evaluation.}
    With a third truth value, errors hidden in binary metrics become visible.
    The increase in \texttt{unknown} predictions highlights the added difficulty of three-valued semantics.
\end{inparaenum}

\subsubsection{Answer Set Verification}
\begin{figure}[h!]
    \centering
    \includegraphics[width=1\linewidth]{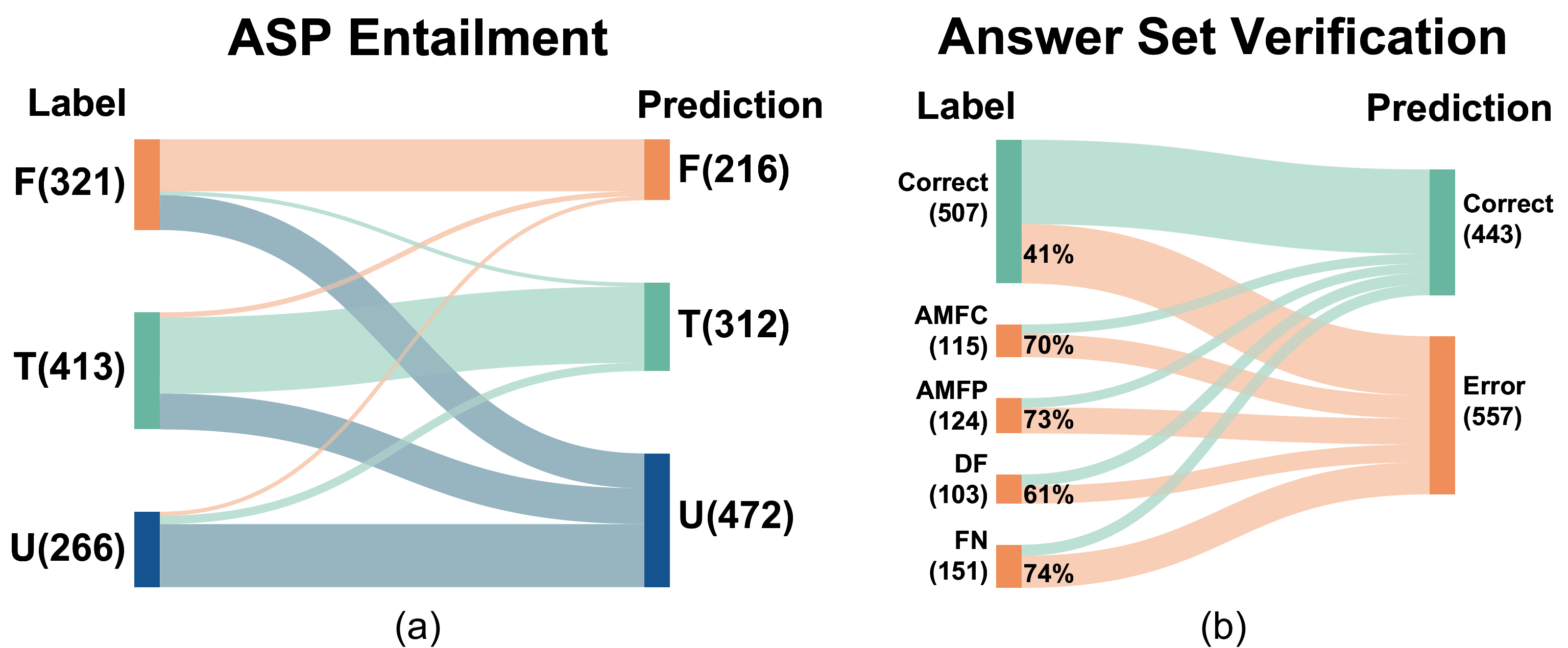}
    \caption{\label{fig:sankey_ASE_and_ASV} Sankey diagrams of average results on ASE and ASV. Truth-value abbreviations: T (true), F (false), U (unknown). Perturbation categories: AMFC (Add Modified Fact -- Constants), AMFP (Add Modified Fact -- Predicate), DF (Delete Fact), and FN (Flip Negation).}
\end{figure}

\begin{figure}[h!]
    \centering
    \includegraphics[width=1\linewidth]{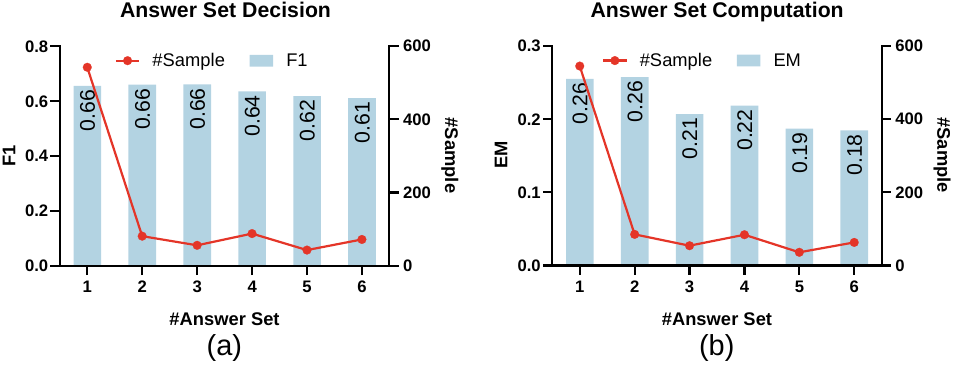}
    \caption{\label{fig:answer_set_num_analysis} Effect of the number of answer sets on performance of LLMs.}
\end{figure}

For answer set verification (Figure~\ref{fig:sankey_ASE_and_ASV} (b) and Figure~\ref{fig:answer_set_num_analysis} (a)), we observe the following insights:

\textbf{Completeness blind spot:}
Ground-truth answer sets (\textit{Correct}) and subsets formed by deleting a single fact (DF) have similar rates of being misjudged (41\% vs.~39\%), notably higher than for other situations.
This indicates LLMs struggle to judge overall consistency, though they perform relatively well on local consistency.

\textbf{Low sensitivity of solution space complexity:}
The performance of the LLM when evaluating candidate answer sets is not significantly affected by the complexity of the program, as measured by the total number of answer sets.
For instance, accuracy only slightly decreases from 66\% for programs with a single answer set to 61\% for those with six.
This general insensitivity indicates that the size of the solution space has a limited impact on this evaluation task.
However, the observed minor decline suggests that very high program complexity could still pose an indirect challenge to these localized consistency checks.

\subsubsection{Answer Set Computation}
For answer set computation (Figure~\ref{fig:answer_set_num_analysis} (b)), we observe that the performance of LLMs is sensitive to the number of answer sets in the program. 
Compared to ASV, LLMs are more sensitive to the number of answer sets in ASC.
The following insights are observed:

\textbf{Collapse in Model Computation:} 
ASC requires the construction of a complete and coherent answer set based on the ASP program. 
This is a far more complex process than ASV, which, related to model checking, primarily involves verifying whether a given candidate is a valid answer set.
This difference in complexity is clearly reflected in the performance of LLMs, with their ability to perform ASC tasks decreasing sharply when ASP programs define three or more answer sets (e.g., EM from 26\% for two answer sets to 21\% for~three).

\subsection{Test-time and Model Scaling}
\label{sec:test_time_and_model_scaling}
\begin{figure}[htp]
    \centering
    \includegraphics[width=.75\linewidth]{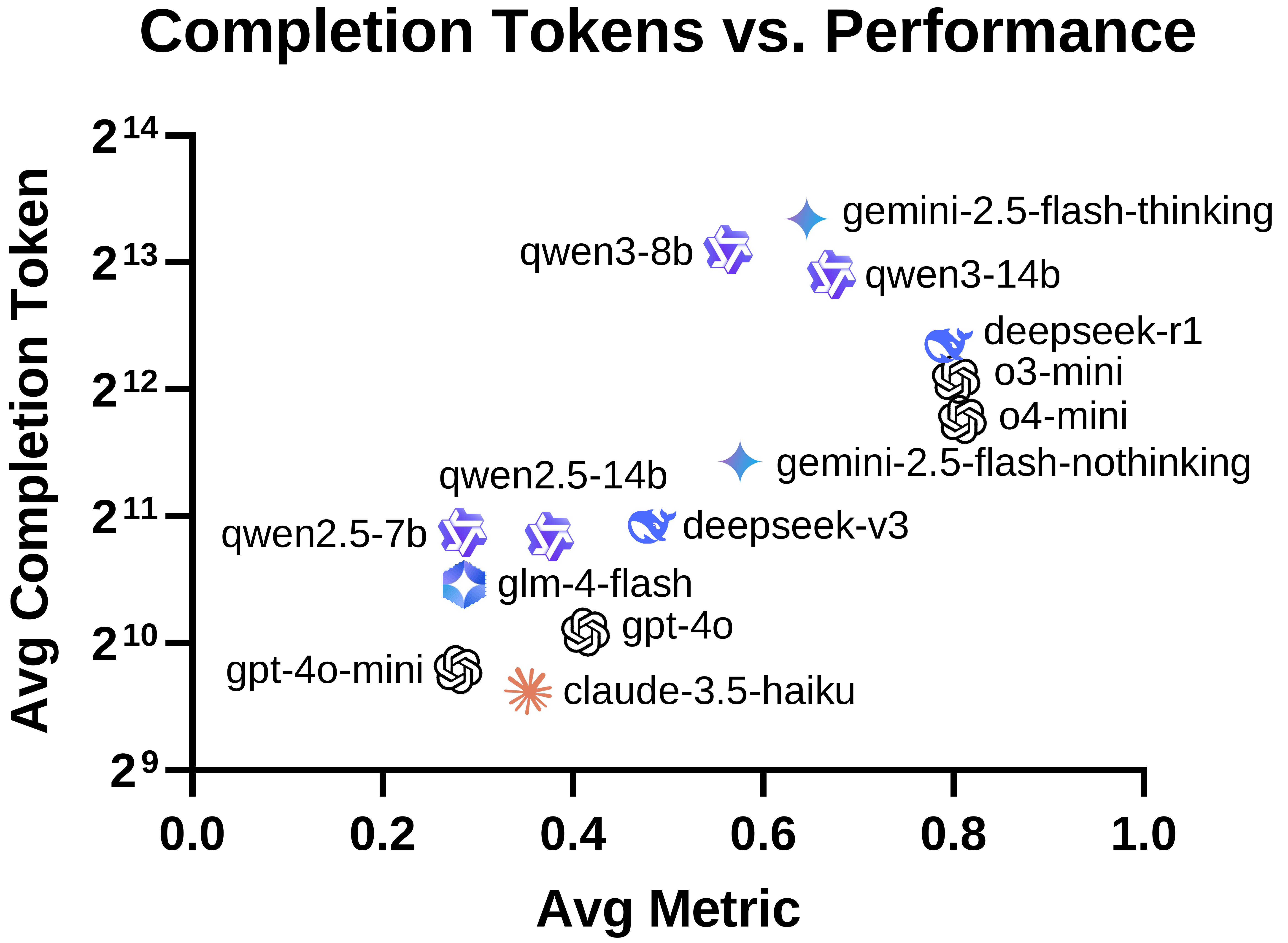}
    \caption{Average completion tokens vs. mean performance of LLMs. Mean performance combines metrics from ASP entailment (F1), answer set verification (F1), and answer set computation (EM).}
    \label{fig:token_vs_performance}
\end{figure}

Test-time scaling~\cite{muennighoff2025s1} and model scaling~\cite{kaplan2020scaling} are two approaches to improve the performance of LLMs. 
To analyze the effect of these two approaches in our tasks, we report the average completion tokens and mean performance of LLMs, as shown in Figure~\ref{fig:token_vs_performance}.

The results visualize the link between average completion tokens (a proxy for thinking depth) and overall score:
\begin{inparaenum}[\itshape (1)]
    \item \textbf{Longer reasoning chains tend to result in higher performance.} 
    Reasoning-oriented variants such as \emph{deepseek-r1}, \emph{gemini-2.5-flash-thinking}, and \emph{qwen3-14b} write longer chains than equally-sized base models and, in return, achieve noticeably higher performance—evidence that letting a model "think longer" at test time pays off. 
    \item \textbf{Larger models are more token-efficient.} 
    Increasing parameters within the same family (e.g., \emph{qwen3-14b} vs. \emph{qwen3-8b}, \emph{gpt-4o} vs. \emph{gpt-4o-mini}) lifts performance while keeping chain length almost unchanged, showing that larger capacity delivers more signal per token.
\end{inparaenum}

Overall, this suggests that when GPU memory is limited, extending the reasoning chain is a cost-effective boost; with sufficient 
resources, scaling model parameters yield more reliable and shorter answers.

\subsection{Performance on Different Syntactic Classes}
\begin{figure}[h!]
    \centering
    \includegraphics[width=1\linewidth]{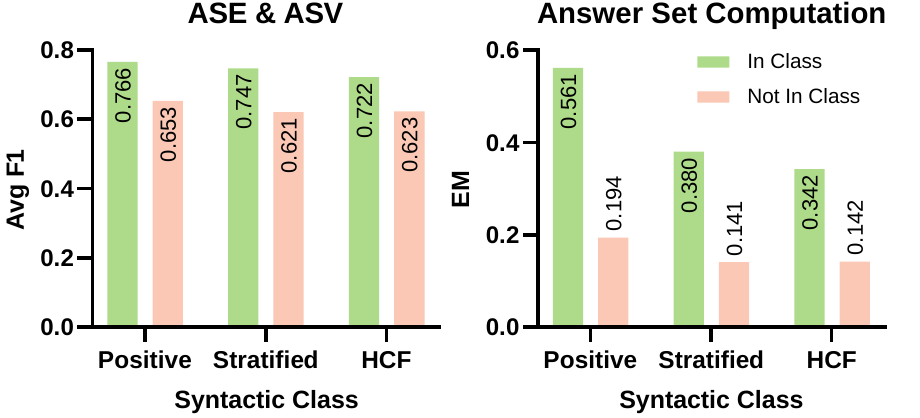}
    \caption{The fine-grained statistic of performance of LLMs on ASPBench with different syntactic classes.} % in class, not in class
    \label{fig:overall_logic_analysis}
\end{figure}

The syntactic structure of logic programs profoundly influences LLM reasoning performance (Figure \ref{fig:overall_logic_analysis}).
When programs adhere to specific syntactic constraints—Positive, Stratified, or HCF—LLMs demonstrate a striking improvement.
This is most evident in ASC, where, for example, compared to non-Positive programs, Positive programs achieve a surge in EM scores from 19.4\% to 56.1\%—\textit{nearly a threefold increase}.
In simpler tasks like ASE and ASV, F1 scores are also boosted by over 10\% when the positive constraint is met.

Furthermore, when cycles that violate Stratification or HCF constraints are present in programs, LLM performance also dramatically degrades.
Particularly in ASC, under such conditions, EM scores fall by over 20\% compared to programs that adhere to the respective Stratified or HCF constraints.
Even for ASE and ASV, the presence of such unconstrained cycles causes scores to plateau around 0.60.
This is significantly below the performance observed for programs that are Positive, or adhere to Stratification or HCF constraints.

This sharp divide highlights a fundamental deficit.
While LLMs perform well with simple positive programs, their performance drops sharply when Stratification or HCF constraints are violated.
In such cases, their average scores fall below those for general non-positive programs.
\textbf{This exposes a critical lack of robust iterative and fixed-point reasoning in LLMs.}

\section{Conclusion}
\label{sec:conclusion}
In this work, we introduce ASPBench, a benchmark designed to evaluate ASP solving ability of LLMs. ASPBench includes diverse descriptions, predicates, and a rich set of logical operations. We define three key tasks: ASP entailment, answer set verification, and answer set computation, to rigorously assess LLM performance. Our experiments reveal significant limitations in the current ability of LLMs to handle ASP solving tasks. Here are a few potential future research directions that could mitigate the aforementioned limitations: 
\begin{inparaenum}[\itshape (1)]
    \item Develop hybrid architectures that integrate symbolic logic representation with neural networks to leverage the strengths of both approaches; 
    \item Propose new innovative methods specifically tailored to enhance ASP solving capability in LLMs.
\end{inparaenum}

\section*{Acknowledgments}
This work is partially supported by National Nature Science Foundation of China under No. 62476058. We thank the Big Data Computing Center of Southeast University for providing the facility support on the numerical calculations in this paper.

\bibliographystyle{kr}
\bibliography{kr-sample}

\appendix
\section{Detailed in Dataset Generation}
\subsection{ASP Graph Construction}
\subsubsection{Rule Graph Construction}
The Rule Graph forms the foundational structure of dependencies between rules.
It is generated as a directed acyclic graph (DAG) where a designated node (typically node 0) serves as the sole sink, meaning all paths eventually lead to this node, and it has no outgoing edges.
The construction ensures that every other node has a path to this sink node.
The parameters guiding this process are outlined in Table~\ref{tab:rule_graph_params}.

\begin{table*}[h!]
\centering
\caption{Parameters for Rule Graph Generation}
\label{tab:rule_graph_params}
\begin{tabularx}{\linewidth}{lX}
\toprule
\textbf{Parameter} & \textbf{Description} \\
\midrule
Number of Rule Nodes & Specifies the total number of rule nodes (e.g., $N_R$) in the graph. \\ \hline
Target Number of Edges & Defines the desired total number of directed edges between the rule nodes. \\ \hline
Random Seed & An integer value used to initialize the random number generator, allowing for reproducible graph structures. \\ \bottomrule
\end{tabularx}
\end{table*}

The process for constructing the Rule Graph unfolds as follows:

\begin{enumerate}
    \item \textbf{Initialization and Validation}:
    \begin{itemize}
        \item The random number generator is initialized using the provided Random Seed, if specified.
        \item Basic validation is performed. For instance, if only one rule node is specified, no edges can be formed.
        \item The requested Target Number of Edges is checked against the minimum and maximum possible edges for a DAG with the given Number of Rule Nodes and the single-sink constraint.
        The minimum number of edges required is (Number of Rule Nodes - 1) to ensure all nodes can reach the sink.
        The maximum number of edges considers all possible non-cyclic connections where the sink node has no outgoing edges.
        If the target number is outside this feasible range, an error is indicated.
        \item An adjacency matrix is initialized to represent the graph, with no edges initially.
    \end{itemize}

    \item \textbf{Constructing Core Connectivity (Ensuring Reachability to Sink Node)}:
    This step ensures that every rule node (other than the sink itself) has a directed path leading to the designated sink node (node 0).
    This effectively creates a spanning in-arborescence directed towards the sink.
    \begin{itemize}
        \item The sink node (node 0) is considered initially "connected" to the structure.
        \item All other rule nodes (from 1 to Number of Rule Nodes - 1) are processed in a random order.
        \item For each such rule node, a "parent" node is randomly selected from the set of nodes already connected to the sink-oriented structure (initially, only the sink node itself is in this set).
        \item A directed edge is added from the current rule node to its selected "parent" node.
        \item The current rule node is then added to the set of "connected" nodes.
        \item This process generates (Number of Rule Nodes - 1) edges and guarantees that all nodes can reach the sink node 0 without forming cycles.
        This can result in various topologies, such as star-like (all nodes directly connect to the sink), chain-like, or more complex branching structures, depending on the random parent selections.
    \end{itemize}
    If the Target Number of Edges equals (Number of Rule Nodes - 1), the graph construction is complete at this stage.

    \item \textbf{Adding Supplementary Edges (Achieving Target Edge Count)}:
    If the Target Number of Edges is greater than the (Number of Rule Nodes - 1) edges created in the previous step, additional edges are incorporated.
    \begin{itemize}
        \item A topological sort of the currently constructed graph is performed.
        This ordering is crucial for adding new edges without introducing cycles.
        \item A list of all possible candidate edges $(u, v)$ is generated.
        An edge is a candidate if:
        \begin{inparaenum}[\itshape (a)]
            \item node $u$ is not the sink node (node 0, which cannot have outgoing edges);
            \item node $u$ appears before node $v$ in the topological sort order (ensuring acyclicity); and
            \item the edge $(u, v)$ does not already exist in the graph.
        \end{inparaenum}
        \item The list of candidate edges is randomly shuffled.
        \item The required number of additional edges (Target Number of Edges minus edges already used) is selected from the top of the shuffled candidate list.
        These selected edges are then added to the graph's adjacency matrix.
    \end{itemize}

    \item \textbf{Resulting Rule Graph}:
    The final output is the adjacency matrix.
    This matrix represents the Rule Graph, a DAG where node 0 is the sole sink, all other nodes can reach node 0, and the graph contains the Target Number of Edges.
    An edge from rule $N_R^i$ to rule $N_R^j$ in this graph signifies a dependency, typically interpreted as the head predicate of $N_R^i$ appearing in the body of $N_R^j$.
\end{enumerate} 

\
\subsubsection{Rule-Predicate Graph Construction}
The expansion from a Rule Graph to a Rule-Predicate Graph is governed by parameters that control the introduction of new predicate nodes and the establishment of additional connections, as detailed in Table~\ref{tab:rule_predicate_graph_params}. These parameters allow for fine-tuning the complexity and structure of the resulting bipartite graph.

\begin{table*}[h!]
\centering
\caption{Parameters for Rule-Predicate Graph Expansion}
\label{tab:rule_predicate_graph_params}
\begin{tabularx}{\linewidth}{lX}
\toprule
\textbf{Parameter} & \textbf{Description} \\
\midrule
Number of Extra Predicates & Specifies the quantity of additional predicate nodes to be created beyond those introduced during the basic expansion. \\ \hline
Number of Extra Edges & Defines the target number of additional ``predicate $\rightarrow$ rule'' edges to be incorporated into the graph. \\ \hline
Max Predicates per Rule Body & Imposes a constraint on the maximum number of distinct predicates that can form the body (antecedent/premises) of any single rule. \\ \hline
Random Seed & An integer value for initializing the random number generator.
\\ \bottomrule
\end{tabularx}
\end{table*}

Constructing the Rule-Predicate Graph involves transforming an initial graph of interconnected logical rules (the Rule Graph) into a more detailed structure that explicitly includes ``predicates.'' Predicates can be conceptualized as the facts or intermediate conclusions that rules produce or depend upon.

The process unfolds as follows:

\begin{enumerate}
    \item \textbf{Linking Rules with Predicates}:
    \begin{itemize}
        \item When the conclusion of a rule $r_i$ contributes to the derivation of another rule $r_j$ (represented as an edge $r_i \rightarrow r_j$ in the original Rule Graph), we introduce an intermediary predicate $p_{ij}$. This signifies that rule $r_i$ produces predicate $p_{ij}$, which is then used as a condition by rule $r_j$. The original link $r_i \rightarrow r_j$ is thus expanded to $r_i \rightarrow p_{ij} \rightarrow r_j$.
        \item The designated initial rule, $r_0$, is also made to produce its own dedicated predicate, $p_0$, represented by the edge $r_0 \rightarrow p_0$.
    \end{itemize}

    \item \textbf{Ensuring Each Rule Has Inputs (Predicates)}:
    A check is performed on every rule to confirm that it uses at least one predicate as a premise for its derivation. If a rule $r_k$ is found to have no incoming predicate dependencies (i.e., no $p \rightarrow r_k$ edges), a new predicate $p_{\text{new}}$ is introduced, and an edge $p_{\text{new}} \rightarrow r_k$ is added. This step ensures that no rule operates in a vacuum and each has defined ``ingredients.''

    \item \textbf{Optional Enhancements for Complexity}:
    The graph can be further elaborated based on specific needs for complexity or density, utilizing parameters such as those listed in Table~\ref{tab:rule_predicate_graph_params}:
    \begin{itemize}
        \item \textit{Adding More Predicates}: If desired, a specified quantity of additional, new predicate nodes (see ``Number of Extra Predicates'' in Table~\ref{tab:rule_predicate_graph_params}) can be created and integrated into the graph.
        \item \textit{Adding More Connections}: Similarly, a specified number of additional connections (see ``Number of Extra Edges'' in Table~\ref{tab:rule_predicate_graph_params}) can be established between existing or newly added predicates and rules. This means a predicate might be used by more rules, or a rule might use more predicates as premises. However, these additions are constrained:
        \begin{itemize}
            \item There is a predefined limit on how many distinct predicates a single rule can use as its inputs or premises (see ``Max Predicates per Rule Body'' in Table~\ref{tab:rule_predicate_graph_params}).
            \item If new predicates are introduced, the system attempts to ensure each is used by at least one rule, making them relevant to the logical structure.
            \item New connections must not create cyclical dependencies in the graph (e.g., where rule $r_i$ depends on predicate $p_k$, which is produced by rule $r_j$, which in turn depends on predicate $p_l$ produced by rule $r_i$), as this would represent a logical fallacy.
        \end{itemize}
    \end{itemize}

    \item \textbf{Final Refinement}:
    Finally, any placeholder nodes or space allocated during the construction process that were not ultimately utilized are removed. This results in the clean, final Rule-Predicate Graph. A record is also kept that distinguishes each node as either a ``rule'' or a ``predicate.''
\end{enumerate}

\subsubsection{Rule-Predicate-Operation Graph Construction}
The Rule-Predicate-Operation Graph is the final stage in our graph generation, building upon the Rule-Predicate Graph by incorporating specific Answer Set Programming (ASP) operators: strong negation (often denoted as \texttt{neg} or ``-'') and default negation (denoted as \texttt{not}). This step applies these negations probabilistically to predicate nodes and reflects these operations on the graph edges. The parameters guiding this process are outlined in Table~\ref{tab:negation_params}.

\begin{table*}[h!]
\centering
\caption{Parameters for Negation Application (Rule-Predicate-Operation Graph)}
\label{tab:negation_params}
\begin{tabularx}{\linewidth}{lX}
\toprule
\textbf{Parameter} & \textbf{Description} \\
\midrule
Strong Negation Probability & Specifies the probability that strong negation will be applied to any given predicate node. \\ \hline
Default Negation Probability & Specifies the probability that default negation will be applied to any given predicate node. \\ \hline
Random Seed & An integer value used to initialize the random number generator for reproducible negation assignment. \\ \bottomrule
\end{tabularx}
\end{table*}

The transformation involves the following conceptual steps:

\begin{enumerate}
    \item \textbf{Probabilistic Assignment of Negations to Predicates}: Each predicate node identified in the Rule-Predicate Graph is considered. Based on the specified probabilities (see Table~\ref{tab:negation_params}), a predicate may be marked to have strong negation, default negation, or potentially both, associated with it.

    \item \textbf{Propagating Negations to Edges}: When a predicate node is assigned a negation, this logical operation is embedded into the properties of all edges connected to it. This includes:
    \begin{itemize}
        \item Edges incoming to the predicate (e.g., $r \rightarrow p$): If rule $r$ defines predicate $p$, and $p$ is now, for instance, strongly negated, this edge now signifies that $r$ defines the strong negation of $p$.
        \item Edges outgoing from the predicate (e.g., $p \rightarrow r$): If predicate $p$ is part of the body of rule $r$, and $p$ is now, for instance, under default negation, this edge signifies that the default negation of $p$ is a premise in rule $r$.
    \end{itemize}
    This is typically achieved by modifying the value or weight of the edge in the graph's adjacency matrix to encode the presence and type of negation (e.g., using distinct bitwise flags for strong and default negation).

    \item \textbf{Resulting Graph}: The outcome is a new adjacency matrix representing the Rule-Predicate-Operation Graph. The edge values in this matrix now carry richer semantic information, indicating not just the basic dependency or definition relationship, but also whether these relationships involve negated predicates, and the type of negation applied.
\end{enumerate}

This enriched graph provides a more detailed structural representation of an ASP-like program, capturing how negations modulate the interactions between rules and predicates.

\section{Examples of Textual ASP Facts and Rules}

Table~\ref{tab:asp_examples} shows some representative examples of the textual ASP facts and rules.

\begin{table*}[htbp] % Use table* for full width
  \centering
  \caption{Representative ASP Facts and Rules with Textual Descriptions} % Updated caption
  \label{tab:asp_examples}
  \small % Use smaller font for the table
  \resizebox{\textwidth}{!}{% Add resizebox here
  \begin{tabular}{lll} % Changed to three columns
    \toprule
    Type & Symbolic Example & Textual Description \\\\ % Added third column header
    \midrule
    Fact (Positive) & \texttt{kissing\_and\_drinking("David", "Jason").} & kissing\_and\_drinking(David, Jason) is true. \\\\
    \midrule
    Fact (Negative) & \texttt{-learning\_about\_science("David", "Lori").} & learning\_about\_science(David, Lori) is explicitly false. \\\\
    \midrule
    Rule (Default \& Strong Neg.) & \parbox[t]{6cm}{\texttt{white(V1, v2) :-  holds(V1, v2),  not fluffy(V1).}} & \parbox[t]{8cm}{If holds(V1, v2) is true and there is no evidence that fluffy(V1) is true, then white(V1, v2) is true.} \\\\
    \bottomrule
  \end{tabular}
  } % Closing brace for resizebox
\end{table*} % Use table* for full width  

\section{Detailed Prompts for LLMs}
\subsection{Prompt for Textualization}
The Detailed Prompt for Textualization is shown in Table \ref{tab:textualization_prompt}. To ensure the quality of the output, we provide detailed examples for various situations.

\subsection{Prompt for ASP Entailment}
The Detailed Prompt for ASP Entailment is shown in Table \ref{tab:asp_entailment_prompt}.

\subsection{Prompt for Answer Set Verification}
The Detailed Prompt for Answer Set Verification is shown in Table \ref{tab:answer_set_verification_prompt}.

\subsection{Prompt for Answer Set Computation}
The Detailed Prompt for Answer Set Computation is shown in Table \ref{tab:answer_set_computation_prompt}.

\section{An Example of Dlv2}
\label{app:syntax_of_dlv2}
The symbolic sample of the scenario is as follows:
\begin{footnotesize}
\begin{lstlisting}[language=prolog]
Bird("Tweety").
Injured("Tweety").
SlightlyInjured("Tweety").
CanFly(A) :- Bird(A), not Abnormal(A).
Abnormal(A) :- Injured(A), 
    not SlightlyInjured(A).
\end{lstlisting}
\end{footnotesize} 

\section{Rule Cover of ASPBench in ASP}
\label{app:Rule_Cover_of_ASPBench_in_ASP}
We have summarized and listed the ASP's constructs in Table~\ref{tab::ASP_Common_constructs}. Our dataset covers most of the constructs of ASP programs and we support all the core features of ASP (``Negation as Failure" and "Disjunctive Rules"). Note that the constructs we do not support all belong to ASP extension extensions or syntax sugar.

\begin{table*}[]
\centering

\resizebox{\textwidth}{!}{
\resizebox{1.0\linewidth}{!}{\setlength{\tabcolsep}{0.1cm} \renewcommand{\arraystretch}{1.0}\begin{tabular}{p{2cm}|p{10cm}|p{5cm}|c}
\specialrule{1pt}{1pt}{1pt}
Construct & Explanation & Example & ASPBench \\ \specialrule{0.5pt}{1pt}{1pt}
Atoms & Basic facts or entities in the domain. & bird(sparrow) & $\surd$ \\ \specialrule{0.5pt}{1pt}{1pt}
Literals & An atom or its negation. & fly(sparrow) or - fly(sparrow) & $\surd$ \\ \specialrule{0.5pt}{1pt}{1pt}
Rules & Implications that define relationships between atoms (head :- body). & fly(X) :- bird(X), - penguin(X). & $\surd$ \\ \specialrule{0.5pt}{1pt}{1pt}
Facts & Ground rules with no body, representing axioms. & bird(sparrow). & $\surd$ \\ \specialrule{0.5pt}{1pt}{1pt}
Constraints & Rules without heads, used to restrict valid solutions. & :- fly(X), penguin(X). & $\surd$ \\ \specialrule{0.5pt}{1pt}{1pt}
Choice Rules & Rules defining optional inclusion of atoms in answer sets. & \{fly(X)\} :- bird(X). &  \\ \specialrule{0.5pt}{1pt}{1pt}
Cardinality Constraints & Bounds on the number of satisfied literals. & 1 \{ fly(X) : bird(X) \} 2. &  \\ \specialrule{0.5pt}{1pt}{1pt}
Aggregates & Functions (sum, count, min, max) applied to collections of literals. & totalWeight(W) :- W = \#sum \{ weight(X) : selected(X) \}. &  \\ \specialrule{0.5pt}{1pt}{1pt}
Negation as Failure & True if a   literal cannot be proven true (negation by failure). & safe(X) :- not   unsafe(X). & $\surd$ \\ \specialrule{0.5pt}{1pt}{1pt}
Strong Negation & Classical   negation, explicitly denoted by -. & -fly(X) :- penguin(X). & $\surd$ \\ \specialrule{0.5pt}{1pt}{1pt}
Disjunctive Rules & Rules with multiple possible outcomes (disjunction in the head). & fly(X) $|$ swim(X) :- bird(X). & $\surd$ \\ \specialrule{0.5pt}{1pt}{1pt}
Optimization Statements & Used to minimize or maximize an objective function. & \#minimize \{   cost(X): selected(X) \}. &  \\ \specialrule{1pt}{1pt}{1pt}
\end{tabular}}

}
\caption{
Common constructs of ASP programs. \label{tab::ASP_Common_constructs}
}
\end{table*} 

\section{Case Studies}
\subsection{ASP Entailment}

An example of ASP entailment is shwon in Table~\ref{tab:case_study_ASE}.

\textbf{LLM Performance Insights:}

\begin{itemize}
    \item \textbf{\textit{o4-mini:}} Successfully deduced the correct \texttt{False} state.

    \item \textbf{\textit{deepseek-r1} and \textit{qwen3-14b:}} Both models incorrectly concluded \texttt{Unknown}. Their failures stemmed from a common root: an inability to correctly derive crucial intermediate predicates necessary for the negation \texttt{-coptic\_church("Savannah","Matthew")}.
    Specifically, both models faltered in deriving \texttt{deadwood(V2)}. This was because they failed to establish its prerequisite, \texttt{rocks(V0,V1)}.
    The error of \textit{deepseek-r1} can be traced further back to an incorrect attempt to derive \texttt{-preamble("Scott")} (which was impossible) instead of the correct and derivable \texttt{-preamble("Matthew")}, which is a necessary precursor for \texttt{rocks("Matthew","Matthew")}. \textit{qwen3-14b} similarly reported that "\texttt{rocks(V0,V1)} is not derived," halting the deductive path.
\end{itemize}

This case underscores that errors in ASP reasoning for LLMs often arise from incorrect variable unification or a failure to explore the correct derivation path for intermediate sub-goals. The inability to derive \texttt{rocks} and subsequently \texttt{deadwood} was the direct cause of the "Unknown" responses from the failing models.

\subsection{Answer Set Verification}
An example of answer set verification is shwon in Table~\ref{tab:case_study_ASV}.

\textbf{LLM Performance Insights:}

\begin{itemize}
    \item \textbf{\textit{o4-mini:}} Correctly determined that the candidate set is a valid answer set. 

    \item \textbf{\textit{o3-mini:}} Incorrectly concluded that the candidate set was not valid. This error stemmed from a form of hallucination: it misquoted Rule 6 from the input program, specifically by altering the sign of its head literal from negative (\texttt{-winter\_sport}) to positive (\texttt{winter\_sport}). This fundamental misrepresentation of the rule led to an erroneous stability check.

    \item \textbf{\textit{gemini-2.5-flash-thinking:}} Also incorrectly found the candidate set to be invalid due to a fundamental misunderstanding of the Gelfond-Lifschitz reduct construction. It mistakenly excluded rules from the reduct (e.g., Rule 2, \texttt{-depth(V3,V3) :- -fraternization(V3).}) if their bodies were not immediately satisfied by the candidate answer set, even when those rules contained no default negations. Such a flawed reduct construction naturally led to an incorrect assessment.
\end{itemize}

This case highlights that LLM errors in ASV can arise from both misrepresentation of program rules (hallucination) and incorrect application of core ASP theoretical constructs like the Gelfond-Lifschitz reduct.

\subsection{Answer Set Computation}
An example of answer set computation is shwon in Table~\ref{tab:case_study_ASC}.

\textbf{LLM Performance Insights:}

\begin{itemize}
    \item \textbf{\textit{deepseek-r1:}} Successfully computed the correct and complete answer set. It correctly identified all facts and derived all necessary literals through iterative rule application.

    \item \textbf{\textit{o4-mini:}} Provided a partially correct answer set but failed to derive all literals. Specifically, it missed \texttt{P1("Christopher")}, \texttt{-P0("Christopher", "Charles")}, and \texttt{-P6("Christopher", "Christopher", "Christopher")} (highlighted in blue in its output within Table~\ref{tab:case_study_ASC}). This indicates an incomplete application of rules of the program or an error in evaluating default negation for certain predicates, leading to an incomplete model. It incorrectly claimed that "all and only these literals are derivable."

    \item \textbf{\textit{qwen3-14b:}} Failed to produce a correct answer set due to multiple errors.
    Firstly, it made an incorrect derivation, concluding \texttt{-P4("Barbara", "John", "Charles")} instead of the correct \texttt{-P4("Barbara", "Christopher", "Charles")} (Rule 8 error highlighted in red).
    Secondly, it demonstrated a misunderstanding of default negation (NAF), stating for Rule 15 that if a predicate like \texttt{P18("John", "Barbara", "Lucas")} is not mentioned in facts, "unless there is a rule that derives it, we can't assume it is true or false." This is incorrect under ASP semantics, where failure to derive implies falsity for the purpose of \texttt{not}. This misunderstanding prevented it from deriving \texttt{P10("Lucas", "Lucas", "Lucas")}.
    Consequently, its final answer set was missing several literals (e.g., \texttt{P16("Christopher", "Charles", "Lucas")}, \texttt{P2("Christopher", "Charles", "Lucas")}, and \texttt{P10("Lucas", "Lucas", "Lucas")}, highlighted in orange in output of \textit{deepseek-r1} for comparison) and included an incorrectly derived literal.
\end{itemize}

This case study demonstrates that while some LLMs can correctly compute answer sets, others struggle with the iterative nature of ASP, correct variable unification, and the precise semantics of default negation. Errors in these areas can lead to incomplete or incorrect answer sets.

\onecolumn % Switch to one-column mode
\begin{longtable}{p{\textwidth}}
\caption{Prompt for Textualization}\label{tab:textualization_prompt} \\ % Caption and label go first in longtable
\hline
\# Task: Convert the given Answer Set Programming (ASP) program into clear, unambiguous natural language text. \\

\\
\textbf{\#\# About Answer Set Programming (ASP) and Input Format Transformation} \\

ASP is a declarative programming paradigm. You will receive symbolic items (facts, rules, queries). \\
*   \textbf{Facts (Potentially Transformed):} Represent basic truths, ending with a period (`.'). \\
    *   Positive facts remain unchanged: `bird(tweety).' \\
    *   \textbf{Strongly negated facts are transformed:} Original `-fact(a).' becomes '$<$strong negation$>$ fact(a).` for clarity. \\
*   \textbf{Rules (Transformed Format):} Rules have been pre-processed for clarity. They appear in the format: \\
  `Original Rule. :: [if] Processed Body [then] Processed Head' \\
    *   `Original Rule.' is the standard ASP rule. \\
    *   `::' separates the original rule from its transformed representation. \\
    *   `[if] Processed Body' contains the conditions (rule body). \\
    *   `[then] Processed Head' contains the conclusion (rule head). \\
    *   \textbf{The core logic remains: If the conditions in `[if]' are met, then the conclusion in `[then]' holds.} \\
*   \textbf{Negation Markers:} To avoid ambiguity, negations within the `[if]' and `[then]` parts are explicitly marked: \\
    *   \textbf{`$<$ default negation $>$':} Replaces the standard `not'. It means "it cannot be assumed that..." or "there is no evidence that...". \\
    *   \textbf{`$<$ strong negation $>$':} Replaces the standard `-'. It means "it is explicitly known that ... is not true" or "it is definitively false that...". \\
    *   \textbf{Example Transformation:} \\
        *   Original: `-p(X) :- q(X), not -r(X).' \\
        *   Transformed Input: `-p(X) :- q(X), not -r(X). :: [if] q(X), $<$default negation$>$ $<$strong negation$>$ r(X) [then] $<$strong negation$>$ p(X)' \\
*   \textbf{Literals:} Basic components in the `[if]' and `[then]' parts, possibly preceded by negation markers. \\
*   \textbf{Query:} A question ending with `$?$'. Example: `fly(tweety)$?$' means "Is it true that Tweety flies$?$". Queries are not transformed. \\

\\
\textbf{\#\# Input Format} \\

You will receive the ASP program structured as follows: \\

[facts]: \\
fact 1: ... \\
fact 2: ... \\
... \\

[rules]: \\
rule 1: ... \\
rule 2: ... \\
... \\

[query] \\
... \\

\\
\textbf{\#\# Your Task: Textualization} \\

Translate the provided ASP facts, rules, and query into natural language text, strictly adhering to the following guidelines, paying attention to examples for different predicate arities (0-ary, 1-ary, 2-ary, 2+-ary): \\

1.  \textbf{Express Facts:} Translate each fact into a simple declarative sentence, preserving all arguments and their order. \\
    *   Example (0-ary): `system\_ready.` $\rightarrow$ "The system is ready." \\
    *   Example (0-ary, Strong Negation): `-alarm\_active.` $\rightarrow$ `$<$strong negatio`$>$ alarm\_active.` $\rightarrow$ "alarm\_active is explicitly false." \\
    *   Example (1-ary): `bird(tweety).` $\rightarrow$ "bird(tweety) is true." or "The predicate bird applies to tweety." \\
    *   Example (1-ary, Strong Negation): `-fly(penguin).` $\rightarrow$ `$<$strong negatio`$>$ fly(penguin).` $\rightarrow$ "fly(penguin) is explicitly false." \\
    *   Example (2-ary): `connected(routerA, switchB).` $\rightarrow$ "connected(routerA, switchB) is true." \\
    *   Example (2-ary, Strong Negation): `-parent(adam, cain).` $\rightarrow$ `$<$strong negatio`$>$ parent(adam, cain).` $\rightarrow$ "parent(adam, cain) is explicitly false." \\
    *   Example (2+-ary, e.g., 3-ary): `route(packet1, routerA, port3).` $\rightarrow$ "route(packet1, routerA, port3) is true." \\
    *   Example (2+-ary, Strong Negation): `-owns(personX, carY, year2023).` $\rightarrow$ `$<$strong negatio`$>$ owns(personX, carY, year2023).` $\rightarrow$ "owns(personX, carY, year2023) is explicitly false." \\

2.  \textbf{Express Rules (Using Transformed Format):} Rules are provided in the format `Original Rule. :: [if] Processed Body [then] Processed Head`. Translate them into a conditional statement ("If [body conditions], then [head conclusion]."). \\
    *   Focus on the `[if]` and `[then]` parts for the translation logic. \\
    *   Translate the `Processed Body` as the condition(s) and the `Processed Head` as the conclusion(s). \\
    *   Preserve all predicates, constants, variables, and argument structures exactly as they appear within the `[if]` and `[then]` parts. \\
    *   Translate the explicit negation markers (`$<$default negation$>$`, `$<$strong negation$>$`) according to Guideline 5. \\
    *   \textbf{Example (Rule with Negations):} \\
        *   Original Rule: `-p(X) :- q(X), not -r(X).` \\
        *   Transformed Input: `-p(X) :- q(X), not -r(X). :: [if] q(X),$<$default negatio$>$$<$strong negatio$>$ r(X) [then]$<$strong negatio$>$ p(X)` \\ 
        *   Translation: "If q(X) holds and there is no evidence that r(X) is explicitly false, then p(X) is explicitly false." \\
    *   \textbf{Disjunction in Head:} If the `[then]` part contains `|` (disjunction), translate it as "either ... or ...". Example: `a(X) | b(X) :- c(X). :: [if] c(X) [then] a(X) | b(X)` $\rightarrow$ "If c(X) is true, then either a(X) is true or b(X) is true." \\
    *   \textbf{Integrity Constraints:} These rules have an empty head and signify conditions that cannot hold true simultaneously. They are represented as `:- Body. :: [if] Processed Body [then]$<$fals$>$`. \\
        *   Example: `:- owns(Person, car), minor(Person). :: [if] owns(Person, car), minor(Person) [then]$<$fals$>$` \\
        *   Translation: "It cannot be the case that both owns(Person, car) is true and minor(Person) is true." or "It is impossible for a minor person to own a car." \\
    *   \textbf{Comparison Constraints:} Comparisons (like $>$`, $<$`, `=`, `!=`, $>$=`, $<$=`) can appear in the rule body. They are translated directly. \\
        *   Example: `high\_temp(Sensor) :- temp(Sensor, Value), Value$>$ 100. :: [if] temp(Sensor, Value), Value$>$ 100 [then] high\_temp(Sensor)` \\
        *   Translation: "If temp(Sensor, Value) is true and Value is greater than 100, then high\_temp(Sensor) is true."

3.  \textbf{Preserve Predicates, Constants, and Arity:} Use the \textbf{exact same} predicate names, constant names, and variable names from the symbolic code (found within facts or the `[if]` / `[then]` parts of rules). \textbf{Do not change them or use synonyms.} Ensure \textbf{all} arguments appear in the translation in the \textbf{exact same order}. This applies \textbf{even if arguments are identical}. The structure and number of arguments (arity) must be preserved. \\
    *   Example (0-ary): `error.` must be translated reflecting 'error', e.g., "An error occurred." \\
    *   Example (1-ary): `active(process1).` $\rightarrow$ "active(process1) is true." \\
    *   Example (2-ary, repetition): `relation(A, A).` $\rightarrow$ "relation(A, A) is true." \\
    *   Example (2-ary, strong negation, repetition): `-connected(PortA, PortA).` $\rightarrow$ `$<$strong negation$>$ connected(PortA, PortA).` $\rightarrow$ "connected(PortA, PortA) is explicitly false." \\
    *   Example (2+-ary, e.g., 3-ary): `transfer(AccountX, AccountY, Amount100).` $\rightarrow$ "transfer(AccountX, AccountY, Amount100) is true." \\
    *   Example (2+-ary, e.g., 3-ary, repetition): `between(PointA, PointB, PointA).` $\rightarrow$ "between(PointA, PointB, PointA) is true." \\
    *   \textbf{Correct Examples (Emphasis on Argument Preservation):} \\
        *   `relation(A, A).` must be translated preserving both arguments, like "relation(A, A) is true." Merging them ("A relates to itself") is \textbf{wrong}. \\
        *   `-connected(PortA, PortA).` must be translated preserving both arguments, like `-connected(PortA, PortA).` $\rightarrow$ `$<$strong negation$>$ connected(PortA, PortA).` $\rightarrow$ "connected(PortA, PortA) is explicitly false." Omitting one ("PortA is not connected itself.") is \textbf{wrong}. \\
        *   `between(PointA, PointB, PointA).` must be translated preserving all arguments in order, like "between(PointA, PointB, PointA) is true." Changing order or structure ("PointA is between PointB.") is \textbf{wrong}. \\
    *   \textbf{General Incorrect Practices:} Omitting arguments, merging arguments (e.g., `loves(R, R)` $\rightarrow$ "R loves himself"), or changing order/structure are \textbf{wrong}. \\

4.  \textbf{Consistency:} Ensure the same predicates and constants are translated \textbf{consistently} throughout the text, regardless of arity. \\
    *   Example (0-ary): If `shutdown.` is translated as "A shutdown occurred.", use this phrasing consistently. \\
    *   Example (1-ary): If `bird(tweety).` is "Tweety is a bird," use "Tweety" and "bird" consistently. \\
    *   Example (2-ary): If `parent(adam, cain).` is "adam is the parent of cain.", use "parent", "adam", "cain" consistently in other translations involving them. \\
    *   Example (2+-ary): If `assign(task1, userA, projectX).` is "task1 is assigned to userA for projectX.", maintain this phrasing style for other `assign` facts/rules. \\

5.  \textbf{Translate Explicit Negation Markers:} Translate the $<$default negatio$>$` and $<$strong negatio$>$` markers consistently, following the `Original $>$ Transformed $>$ Translation` pattern where applicable. \\
    *   \textbf{$<$strong negatio$>$`:} Translate as explicit falsehood. Corresponds to original `-p(...)`. \\
        *   Example (0-ary): `-safe.` $\rightarrow$ $<$strong negatio$>$ safe.` $\rightarrow$ "safe is explicitly false." \\
        *   Example (1-ary): `-fly(penguin).` $\rightarrow$ $<$strong negatio$>$ fly(penguin).` $\rightarrow$ "fly(penguin) is explicitly false." \\
        *   Example (2-ary): `-connected(serverA, serverB).` $\rightarrow$ $<$strong negatio$>$ connected(serverA, serverB).` $\rightarrow$ "connected(serverA, serverB) is explicitly false." \\
        *   Example (2+-ary): `-authorized(userX, actionY, resourceZ).` $\rightarrow$ $<$strong negatio$>$ authorized(userX, actionY, resourceZ).` $\rightarrow$ "authorized(userX, actionY, resourceZ) is explicitly false." \\
        *   Use phrases like "\textbf{... is explicitly false}". \\
    *   \textbf{$<$default negatio$>$`:} Translate as failure to prove / lack of evidence. Corresponds to original `not p(...)`. \\
        *   Example (in rule body): `... :- ..., not error\_detected.` $\rightarrow$ `... :: [if] ...,$<$default negatio$>$ error\_detected [then] ...` $\rightarrow$ "If ... and there is no evidence that error\_detected is true, then ..." \\
        *   Example (in rule body): `fly(X) :- bird(X), not broken\_wing(X).` $\rightarrow$ `fly(X) :- bird(X), not broken\_wing(X). :: [if] bird(X),$<$default negatio$>$ broken\_wing(X) [then] fly(X)` $\rightarrow$ "If bird(X) is true and there is no evidence that broken\_wing(X) is true, then fly(X) is true." \\
        *   Example (in rule body): `eligible(P, L) :- citizen(P, C), not has\_loan(P, L).` $\rightarrow$ `eligible(P, L) :- citizen(P, C), not has\_loan(P, L). :: [if] citizen(P, C),$<$default negatio$>$ has\_loan(P, L) [then] eligible(P, L)` $\rightarrow$ "If citizen(P, C) is true and there is no evidence that has\_loan(P, L) is true, then eligible(P, L) is true." \\
        *   Example (in rule body with underscore): `can\_assign(...) :- task(T), ..., not assigned(T, \_, \_).` $\rightarrow$ `... :: [if] task(T), ...,$<$default negatio$>$ assigned(T, \_, \_) [then] can\_assign(...)` $\rightarrow$ "... and there is no evidence that assigned(T, \_, \_) is true for any user and project..." \\
        *   Use phrases like "\textbf{there is no evidence that ... is true}". \\
    *   \textbf{Combination ($<$default negatio$>$$<$strong negatio$>$ p(...)`):} Translate as lack of proof for strong negation. This corresponds to the original ASP pattern `not -p(...)`. \\
        *   Example (0-ary): `not -system\_stable.' $\rightarrow$ $<$default negatio$>$$<$strong negatio$>$ system\_stable.' $\rightarrow$ "There is no evidence that system\_stable is explicitly false."
        *   Example (1-ary): `not -blocked(userX).' $\rightarrow$ $<$default negatio$>$$<$strong negatio$>$ blocked(userX).' $\rightarrow$ "There is no evidence that blocked(userX) is explicitly false."
        *   Example (2-ary): `not -married(personA, personB).' $\rightarrow$ $<$default negatio$>$$<$strong negatio$>$ married(personA, personB).' $\rightarrow$ "There is no evidence that married(personA, personB) is explicitly false." \\
        *   Example (2+-ary): `not -conflict(meeting1, roomA, timeT).' $\rightarrow$ $<$default negatio$>$$<$strong negatio$>$ conflict(meeting1, roomA, timeT).' $\rightarrow$ "There is no evidence that conflict(meeting1, roomA, timeT) is explicitly false." \\

6.  \textbf{Reduce Ambiguity:} Structure sentences clearly. Pay attention to the implicit universal quantification of variables in rules (as seen in the `[if]` and `[then]` parts). \\
    *   Example (2-ary): `ancestor(X, Y) :- parent(X, Y).' $\rightarrow$ ``For any X and Y, if X is the parent of Y, then X is the ancestor of Y." \\
    *   Example (2+-ary): `[if] parent(X, Y), parent(Y, Z) [then] grandparent(X, Z)' $\rightarrow$ ``For any X, Y, and Z, if X is the parent of Y and Y is the parent of Z, then X is the grandparent of Z."

7.  \textbf{Handle the Query:} Queries are provided in their original format (ending with `$?$'). Phrase the query as a question or statement to be verified, preserving its structure and arity. Translate any negations (`-' or `not') directly as they appear in the query. \\  
    *   Example (0-ary): `alarm$?$' $\rightarrow$ Is alarm true$?$" \\
    *   Example (1-ary): `fly(tweety)$?$' $>$ ``Is fly(tweety) true$?$"
    *   Example (2-ary): `connected(routerA, switchB)$?$' $\rightarrow$ ``Is connected(routerA, switchB) true$?$" \\
    *   Example (2-ary, negation): `-parent(adam, cain)$?$' $\rightarrow$ ``Is parent(adam, cain) explicitly false$?$" \\
    *   Example (2+-ary): `route(packet1, routerA, port3)$?$' $\rightarrow$ ``Is route(packet1, routerA, port3) true$?$" \\

8.  \textbf{Rule Generalization:} The transformed rule format already presents the rule generally using variables. Translate it directly based on the `[if]` and `[then]` parts. \\

9.  \textbf{No Direct Copying:} The output must be a natural language translation, not a copy of the symbolic item (whether original or transformed). \\

10. \textbf{Symbolic Nature of Predicates:} Remember predicate names are symbolic labels, regardless of arity (0, 1, 2, 2+). Their meaning comes from the program's logic, not the word itself. Translate them consistently based on the name. \\
    *   Example (0-ary): `flagged.' might just mean a condition is met. Translate as ``flagged is true." \\
    *   Example (1-ary): `red(blockA).' means the predicate 'red' applies to 'blockA'. Translate as ``red(blockA) is true." \\
    *   Example (2-ary): `press(A, B)' represents the symbolic relationship 'press' between A and B. Translate as ``press(A, B) is true." \\
    *   Example (2+-ary, e.g., 3-ary): `link(Node1, Node2, ProtocolA)' represents the symbolic relationship `link' involving Node1, Node2, and ProtocolA. Translate as ``link(Node1, Node2, ProtocolA) is true." \\

\\
\textbf{\#\# Output Format} \\

Present the textualized result clearly. \\

--- \\

Now, textualize the symbolic items listed below. Remember that rules and strongly negated facts are presented in a transformed format using `:: [if]...[then]...` for rules and `$<$strong negation$>$` for facts/rule parts. Translate positive facts and queries directly. Translate transformed items based on the explicit markers (`$<$default negation$>$`, `$<$strong negation$>$`) and rule structure (`[if]...[then]...`). \\

[Symbolic Items to Textualize]: \\
\{\# Render the list of symbolic items passed in the context \#\} \\
\{\% if render\_context.symbolic\_items \%\} \\
  \{\% for item in render\_context.symbolic\_items \%\} \\
    \$ \{\{ item \}\} \$  \\
  \{\% else \%\} \\
    (No symbolic items provided for textualization) \\
  \{\% endfor \%\} \\
\{\% else \%\} \\
(No symbolic items provided in the context) \\
\{\% endif \%\} \\

\hline

\end{longtable}
\twocolumn % Switch back to two-column mode 

\onecolumn % Switch to one-column mode
\begin{longtable}{p{\textwidth}}
\caption{Prompt for ASP Entailment}\label{tab:asp_entailment_prompt} \\ % Caption and label go first in longtable
\hline
\textbf{\# Task:} Determine the truth state of the query based on the given ASP program. \\
\textbf{\#\# Understanding the ASP Program} \\

You are given an Answer Set Programming (ASP) program consisting of facts, rules, and a query. Your goal is to determine if the `[query]` is \textbf{True}, \textbf{False}, or \textbf{Unknown} based on the provided `[facts]` and `[rules]`. \\

*   \textbf{Facts:} Basic statements considered true, ending with a period (`.`). Example: `bird(tweety).` means "bird(tweety) is true." \\
    *   Strongly negated facts like `-fly(penguin).` mean "It is explicitly known that fly(penguin) is false." \\
*   \textbf{Rules:} Define relationships and conclusions in the form `Head :- Body.`. This means "If the `Body` is true, then the `Head` is true." \\
    *   The `Body` can contain literals (like `p(X)`) and negated literals. \\
    *   \textbf{Default Negation (`not`):} `not p(X)` means "It cannot be proven that p(X) is true." or "There is no evidence that p(X) is true." \\
    *   \textbf{Strong Negation (`-`):} `-p(X)` means "It is explicitly known that p(X) is false." This is a stronger claim than `not p(X)`. \\
    *   \textbf{Integrity Constraints:} Rules with an empty head (e.g., `:- p(X), q(X).`) mean the body cannot be true simultaneously. If the body becomes true based on the facts and other rules, it indicates a contradiction in the potential answer set. \\
*   \textbf{Query:} A statement ending with `$?$`. You need to determine its truth value based on the logical consequences of the facts and rules. \\

\#\# How to Determine the Query State \\

1.  \textbf{Apply the rules} to the facts iteratively to derive all possible conclusions (the stable model or answer set). Pay close attention to how `not` and `-` affect the derivability of literals. \\
2.  \textbf{Check the query against the derived conclusions:} \\
    *   \textbf{True:} The query (e.g., `p(a)`) is present in the derived conclusions/stable model. \\
    *   \textbf{False:} The strong negation of the query (e.g., `-p(a)`) is present in the derived conclusions/stable model. \\
    *   \textbf{Unknown:} Neither the query nor its strong negation is present in the derived conclusions/stable model. This typically happens if the information is insufficient to prove either case. \\

\\
\textbf{\#\# Examples} \\

\textbf{Example 1: Query is True} \\

*   \textbf{(Symbolic Program)} \\
    *   Facts: `bird(tweety).` \\
    *   Rules: `fly(X) :- bird(X), not -fly(X).`
    *   Query: `fly(tweety)$?$` \\

*   \textbf{(Textualized Program)} \\
    *   Facts: \\
        *   `bird(tweety)` is true. \\
    *   Rules: \\
        *   If `bird(X)` is true and there is no evidence that `fly(X)` is explicitly false, then `fly(X)` is true. \\
    *   Query: The query is `fly(tweety)`. \\

*   \textbf{(Reasoning)} \\
    1.  We are given the fact "`bird(tweety)` is true" (symbolically `bird(tweety).`). \\
    2.  Consider the rule: "If `bird(X)` is true and there is no evidence that `fly(X)` is explicitly false, then `fly(X)` is true." (symbolically `fly(X) :- bird(X), not -fly(X).`). We apply this for `X = tweety`. \\
    3.  The first condition `bird(tweety)` is true holds (from the fact). \\
    4.  The second condition is "there is no evidence that `fly(tweety)` is explicitly false" (symbolically `not -fly(tweety)`). We check if `-fly(tweety)` ("`fly(tweety)` is explicitly false") can be derived. It cannot be derived from the program. \\
    5.  Therefore, the condition `not -fly(tweety)` holds. \\
    6.  Since both conditions in the rule body are satisfied, the rule allows us to derive the head: "`fly(tweety)` is true" (symbolically `fly(tweety)`). \\

*   \textbf{(Conclusion)} \\
    *   Because `fly(tweety)` ("`fly(tweety)` is true") is derivable and belongs to the stable model, the state of the query `fly(tweety)$?$` is \textbf{True}. \\

\textbf{Example 2: Query is False (Derived)} \\

*   \textbf{(Symbolic Program)} \\
    *   Facts: `penguin(pingu).`, `bird(pingu).` \\
    *   Rules: \\
        *   `fly(X) :- bird(X), not -fly(X).` \\
        *   `-fly(X) :- penguin(X).` \\
    *   Query: `fly(pingu)$?$` \\

*   \textbf{(Textualized Program)} \\
    *   Facts: \\
        *   `penguin(pingu)` is true. \\
        *   `bird(pingu)` is true. \\
    *   Rules: \\
        *   If `bird(X)` is true and there is no evidence that `fly(X)` is explicitly false, then `fly(X)` is true. \\
        *   If `penguin(X)` is true, then `fly(X)` is explicitly false. \\
    *   Query: The query is `fly(pingu)`. \\

*   \textbf{(Reasoning)} \\
    1.  We are given the fact "`penguin(pingu)` is true" (symbolically `penguin(pingu).`). \\
    2.  Consider the second rule: "If `penguin(X)` is true, then `fly(X)` is explicitly false." (symbolically `-fly(X) :- penguin(X).`). \\
    3.  Applying this rule with `X = pingu`, since the condition "`penguin(pingu)` is true" holds, we derive the head: "`fly(pingu)` is explicitly false" (symbolically `-fly(pingu)`). \\

*   \textbf{(Conclusion)} \\
    *   Because `-fly(pingu)` ("`fly(pingu)` is explicitly false") is derivable and belongs to the stable model, the state of the query `fly(pingu)$?$` is \textbf{False}. \\

\textbf{Example 3: Query is Unknown} \\

*   \textbf{(Symbolic Program)} \\
    *   Facts: `bird(polly).` \\
    *   Rules: `fly(X) :- bird(X), can\_fly(X).` \\
    *   Query: `fly(polly)$?$` \\

*   \textbf{(Textualized Program)} \\
    *   Facts: \\
        *   `bird(polly)` is true. \\
    *   Rules: \\
        *   If `bird(X)` is true and `can\_fly(X)` is true, then `fly(X)` is true. \\
    *   Query: The query is `fly(polly)`. \\

*   \textbf{(Reasoning)} \\
    1.  We are given the fact "`bird(polly)` is true" (symbolically `bird(polly).`). \\
    2.  Consider the rule: "If `bird(X)` is true and `can\_fly(X)` is true, then `fly(X)` is true." (symbolically `fly(X) :- bird(X), can\_fly(X).`). We apply this for `X = polly`. \\
    3.  The first condition `bird(polly)` is true holds. \\
    4.  The second condition is "`can\_fly(polly)` is true" (symbolically `can\_fly(polly)`). We check if this can be derived from the program. \\
    5.  There are no facts or other rules that allow us to derive `can\_fly(polly)`. Therefore, the body of the rule cannot be satisfied, and we cannot derive `fly(polly)` using this rule. \\
    6.  We also check if the strong negation `-fly(polly)` ("`fly(polly)` is explicitly false") can be derived. There are no facts or rules allowing this derivation. \\

*   \textbf{(Conclusion)} \\
    *   Since neither `fly(polly)` ("`fly(polly)` is true") nor `-fly(polly)` ("`fly(polly)` is explicitly false") can be derived from the program, the state of the query `fly(polly)$?$` is \textbf{Unknown}. \\

\\
\textbf{\#\# Input Program} \\

[facts]: \\
\{\% for fact in facts \%\} \\
fact \{\{ loop.index \}\} : \{\{ fact \}\} \\
\{\% endfor \%\} \\

[rules]: \\
\{\% for rule in rules \%\} \\
rule \{\{ loop.index \}\} : \{\{ rule \}\} \\
\{\% endfor \%\} \\

[query] \\
\{\{ target\_query \}\} \\

\#\# Your Answer \\

Based on the facts and rules, analyze the logical consequences step-by-step. Explain your reasoning process, showing how the rules are applied to the facts to derive conclusions. Finally, state clearly whether the query is \textbf{True}, \textbf{False}, or \textbf{Unknown}. \\

\textbf{Reasoning:} \\
Explain your step-by-step derivation \\

\textbf{Final Answer:} [True/False/Unknown] \\

\hline

\end{longtable}
\twocolumn % Switch back to two-column mode

\onecolumn % Switch to one-column mode
\begin{longtable}{p{\textwidth}}
\caption{Prompt for Answer Set Verification}\label{tab:answer_set_verification_prompt} \\ % Caption and label go first in longtable
\hline
\# Task: Determine if the Candidate Answer Set is a valid Answer Set (Stable Model) for the given ASP program. \\

\#\# Understanding the Task \\

You are given an Answer Set Programming (ASP) program (consisting of facts and rules) and a specific set of literals called the 'Candidate Answer Set'. Your task is to rigorously check if this candidate set qualifies as a true Answer Set (also known as a Stable Model) for the program. \\

\#\# Key ASP Concepts \\

*   \textbf{Facts:} Ground atoms assumed to be true (e.g., `p(a).`). \\
*   \textbf{Rules:} Statements of the form `Head :- Body.` meaning "If `Body` is true, then `Head` must be true." \\
    *   `Body` can contain positive literals (e.g., `q(X)`), strongly negated literals (e.g., `-r(X)`), and default-negated literals (e.g., `not s(X)`). \\
    *   \textbf{Default Negation (`not`):} `not p` holds if `p` cannot be derived. \\
    *   \textbf{Strong Negation (`-`):} `-p` means `p` is explicitly false. It is a different literal from `p`. \\
*   \textbf{Answer Set (Stable Model):} A set of ground literals `A` is an answer set if it satisfies two main conditions: \\
    1.  \textbf{Consistency:} `A` does not contain a literal `p` and its strong negation `-p` simultaneously. \\
    2.  \textbf{Stability (Gelfond-Lifschitz Reduct):} Let `P` be the ASP program. The reduct `$P^A$` is obtained by:
        a. Removing every rule from `P` where a default-negated literal `not q` appears in the body, and `q` is present in the candidate set `A`. \\
        b. Removing all default-negated literals `not q` from the bodies of the remaining rules. \\
        The candidate set `A` is an answer set if and only if `A` is the *minimal classical model* of the resulting reduct `$P^A$` (which is now a program without default negation). This essentially means that every literal in `A` must be derivable from the facts and the simplified rules (`$P^A$`), and `A` contains nothing more than what is derivable. \\

\#\# Verification Steps \\

To determine if the `[candidate\_answer\_set]` is a valid answer set for the program defined by `[facts]` and `[rules]`: \\

1.  \textbf{Consistency Check:} Verify that the `[candidate\_answer\_set]` does not contain both `p` and `-p` for any ground atom `p`. If it does, it is not a valid answer set. \\
2.  \textbf{Stability Check (using the Reduct concept):} \\
    a.  Construct the reduct `$P^A$` based on the original program `P` and the `[candidate\_answer\_set]` `A`: \\
        i.  Iterate through each rule in `[rules]`. \\
        ii. If a rule contains `not q` in its body and `q` *is* in `A`, discard the rule entirely. \\
        iii. If a rule survives step (ii), remove all `not q` literals from its body. Keep the head and the remaining positive/strongly-negated body literals. \\
    b.  Consider the reduct `$P^A$` (the simplified rules from step 2.a.iii) along with the original `[facts]`. \\
    c.  Determine the minimal classical model of `$P^A$` plus `[facts]`. This means finding the smallest set of ground literals that includes all `[facts]` and is closed under the rules in `$P^A$`. \\
    d.  Compare this minimal model with the `[candidate\_answer\_set]` `A`. They must be identical. If they are, `A` is a stable model. Otherwise, it is not. \\

\#\# Examples \\

\textbf{Example 1: Valid Answer Set} \\

*   \textbf{(Symbolic Program)} `P`: \\
    *   Facts: (None) \\
    *   Rules: `a :- not b.` \\
*   \textbf{(Textualized Program)} \\
    *   Facts: \\
        *   (None) \\
    *   Rules: \\
        *   If there is no evidence that `b` is true, then `a` is true. \\
*   Candidate Answer Set `A`: `\{a\}` \\

1.  \textbf{Consistency Check:} `\{a\}` is consistent (no `p` and `-p`). \textbf{OK.} \\
2.  \textbf{Stability Check:} \\
    a.  Construct Reduct `$P^A$`: \\
        *   Rule `a :- not b.`: Since `b` is *not* in `A = {a}`, the rule is kept, and `not b` is removed. The rule becomes `a :- .` (or simply `a.`). \\
        *   Reduct `$P^A$` is: `a.` \\
    b.  Minimal classical model of `$P^A$`: The smallest set closed under the rule `a.` is `\{a\}`. \\
    c.  Compare: Minimal model `\{a\}` is identical to the candidate set `A = \{a\}`. \textbf{OK.} \\
*   \textbf{Verdict:} Yes, `\{a\}` is a valid answer set. \\

\textbf{Example 2: Invalid Answer Set (Fails Stability)} \\

*   \textbf{(Symbolic Program)} `P`: \\
    *   Facts: (None) \\
    *   Rules: `p :- not q.`, `q :- not p.` \\
*   \textbf{(Textualized Program)} \\
    *   Facts: \\
        *   (None) \\
    *   Rules: \\
        *   If there is no evidence that `q` is true, then `p` is true. \\
        *   If there is no evidence that `p` is true, then `q` is true. \\
*   Candidate Answer Set `A`: `\{p, q\}` \\

1.  \textbf{Consistency Check:} `\{p, q\}` is consistent. \textbf{OK.} \\
2.  \textbf{Stability Check:} \\
    a.  Construct Reduct `$P^A$`: \\
        *   Rule `p :- not q.`: Since `q` *is* in `A = \{p, q\}`, discard this rule. \\
        *   Rule `q :- not p.`: Since `p` *is* in `A = {p, q}`, discard this rule. \\
        *   Reduct `$P^A$` is empty (contains no rules). \\
    b.  Minimal classical model of the empty reduct `$P^A$`: The minimal model is `\{\}` (the empty set). \\
    c.  Compare: Minimal model `\{\}` is \textbf{not} identical to the candidate set `A = \{p, q\}`. \textbf{FAIL.} \\
*   \textbf{Verdict:} No, `\{p, q\}` is not a valid answer set for this program. \\

\#\# Input Format \\

[facts]: \\
\{\% for fact in facts \%\} \\
fact \{\{ loop.index \}\} : \{\{ fact \}\} \\
\{\% endfor \%\} \\

[rules]: \\
\{\% for rule in rules \%\} \\
rule \{\{ loop.index \}\} : \{\{ rule \}\} \\
\{\% endfor \%\} \\

[candidate\_answer\_set]: \\
\{\% for literal in candidate\_answer\_set \%\} \{\{ literal \}\} \{\% if not loop.last \%\}, \{\% endif \%\} \{\% endfor \%\} \\

\#\# Your Answer \\

Perform the Consistency and Stability checks step-by-step based on the definitions above. Explain your reasoning clearly, detailing the construction of the reduct and the comparison of its minimal model with the candidate set. Finally, state clearly whether the `[candidate\_answer\_set]` \textbf{is} or \textbf{is not} a valid answer set.

\textbf{Reasoning:} \\
Explain your step-by-step verification, including consistency check, reduct construction, minimal model derivation, and comparison \\

\textbf{Final Answer:} [Yes/No] \\

\hline

\end{longtable}
\twocolumn % Switch back to two-column mode

\onecolumn % Switch to one-column mode
\begin{longtable}{p{\textwidth}}
\caption{Prompt for Answer Set Computation}\label{tab:answer_set_computation_prompt} \\ % Caption and label go first in longtable
\hline
\# Task: Generate a valid Answer Set (Stable Model) for the given ASP program. \\

\#\# Understanding the Task \\

You are given an Answer Set Programming (ASP) program consisting of facts and rules. Your task is to find *one* valid Answer Set (also known as a Stable Model) for this program. \\

\#\# Key ASP Concepts Recap \\

*   \textbf{Facts:} Ground atoms assumed to be true (e.g., `p(a).`). \\
*   \textbf{Rules:} Statements of the form `Head :- Body.` ("If `Body` is true, `Head` must be true"). \\
    *   `Body` can contain positive literals (`q(X)`), strongly negated literals (`-r(X)`), and default-negated literals (`not s(X)`). \\
    *   \textbf{Default Negation (`not`):} `not p` holds if `p` cannot be derived. \\
    *   \textbf{Strong Negation (`-`):} `-p` means `p` is explicitly false. \\
*   \textbf{Answer Set (Stable Model):} A set of ground literals `A` that is: \\
    1.  \textbf{Consistent:} Does not contain `p` and `-p` simultaneously. \\
    2.  \textbf{Stable:} `A` is the minimal classical model of the program's reduct `$P^A$` (formed by simplifying rules based on `A`). Essentially, everything in `A` must be derivable from the simplified program, and nothing more. \\

\#\# Goal \\

Your goal is to output a single set of ground literals that constitutes a valid Answer Set for the program defined by `[facts]` and `[rules]`. There might be multiple possible Answer Sets; you only need to provide one. \\

\#\# Examples \\

\textbf{Example 1:} \\

*   \textbf{(Symbolic Program)} \\
    *   Facts: `p(a).` \\
    *   Rules: `q(X) :- p(X), not r(X).` \\
*   \textbf{(Textualized Program)} \\
    *   Facts: \\
        *   `p(a)` is true. \\
    *   Rules: \\
        *   If `p(X)` is true and there is no evidence that `r(X)` is true, then `q(X)` is true. \\
*   Input: \\
    *   `[facts]`: `p(a).` \\
    *   `[rules]`: `q(X) :- p(X), not r(X).` \\
*   Possible Output (Generated Answer Set): \\
    *   Symbolic: `\{ p(a), q(a) \}` (Because `p(a)` is a fact, and `r(a)` cannot be derived, so `not r(a)` holds, making `q(a)` derivable). \\
    *   Textualized: `p(a)` is true. `q(a)` is true. \\

\textbf{Example 2:} \\

*   \textbf{(Symbolic Program)} \\
    *   Facts: (None) \\
    *   Rules: `a :- not b.`, `b :- not a.` \\
*   \textbf{(Textualized Program)} \\
    *   Facts: \\
        *   (None) \\
    *   Rules: \\
        *   If there is no evidence that `b` is true, then `a` is true. \\
        *   If there is no evidence that `a` is true, then `b` is true. \\
*   Input: \\
    *   `[facts]`: (None) \\
    *   `[rules]`: `a :- not b.`, `b :- not a.` \\
*   Possible Output (Generated Answer Set): \\
    *   Symbolic: `\{ a \}` (This is one of the two possible stable models. `\{ b \}` is the other. You only need to provide one.) \\
    *   Textualized: `a` is true. \\

\textbf{Example 3:} \\

*   \textbf{(Symbolic Program)} \\
    *   Facts: `p(a).`, `-q(b).` \\
    *   Rules: `r(X) :- p(X), not q(X).` \\
*   \textbf{(Textualized Program)} \\
    *   Facts: \\
        *   `p(a)` is true. \\
        *   `q(b)` is explicitly false. \\
    *   Rules: \\
        *   If `p(X)` is true and there is no evidence that `q(X)` is true, then `r(X)` is true. \\
*   Input: \\
    *   `[facts]`: `p(a).`, `-q(b).` \\
    *   `[rules]`: `r(X) :- p(X), not q(X).` \\
*   Possible Output (Generated Answer Set): \\
    *   Symbolic: `\{ p(a), -q(b), r(a) \}` (Because `p(a)` and `-q(b)` are facts, and `q(a)` cannot be derived, `not q(a)` holds, making `r(a)` derivable). \\
    *   Textualized: `p(a)` is true. `q(b)` is explicitly false. `r(a)` is true. \\

\#\# Input Format \\

[facts]: \\
\{\% for fact in facts \%\} \\
fact \{\{ loop.index \}\} : \{\{ fact \}\} \\
\{\% endfor \%\} \\

[rules]: \\
\{\% for rule in rules \%\} \\
rule \{\{ loop.index \}\} : \{\{ rule \}\} \\
\{\% endfor \%\} \\

\#\# Your Answer \\

Generate *one* set of ground literals that represents a valid Answer Set for the provided ASP program.

\textbf{Generated Answer Set:} \\
\{ [Provide the comma-separated list of literals in the answer set here] \} \\

\hline

\end{longtable}
\twocolumn % Switch back to two-column mode

\onecolumn 
\small
\begin{longtable}{p{\textwidth}}
\caption{Case study for ASP entailment}\label{tab:case_study_ASE} \\
\toprule 
\textbf{\textsc{Input ASP Program:}} \\ \hline
\begin{verbatim}
- squash_ball("Savannah", "David").
- wall("Savannah", "Matthew").
send_rocket_into_space("Savannah", "Scott").
slope("Savannah", "Scott").
- changing("Matthew", "David").
attacker("Matthew", "Scott").
salt_food("Matthew", "David").
- pomegranate("Scott").
- asset("Scott").
dividend_warrant("Scott").
- cow_juice("Matthew", "Scott").
dolphin("Savannah", "Matthew").
- cell("Matthew", "Scott").
snoop("Savannah", "Scott").
feeling_good("Scott", "David").
larid("Savannah", "Scott").
- coptic_church(V0, V1) :- deadwood(V2), arsenic(V0, V1), not stagehand(V1, V2).
deadwood(V2) :- - launching(V2), rocks(V0, V1), not -send_rocket_into_space(V0, V2).
- launching(V2) :- - squash_ball(V0, V3), attacker(V1, V2), not -salt_food(V1, V3).
restaurant_dining_area(V3) :- - squash_ball(V0, V3), 
    attacker(V1, V2), not -salt_food(V1, V3).
arsenic(V0, V1) :- - wall(V0, V1), slope(V0, V2), - changing(V1, V3).
- alarm(V2) :- - wall(V0, V1), slope(V0, V2), - changing(V1, V3).
- stagehand(V2, V2) :- - alarm(V2), not -print_on(V2), not -adult_body(V2).
print_on(V3) :- restaurant_dining_area(V3), arsenic(V0, V1), not changing(V1, V3).
- preamble(V1) :- restaurant_dining_area(V3), arsenic(V0, V1), not changing(V1, V3).
rocks(V2, V2) :- - preamble(V2), not pomegranate(V2), not asset(V2).
adult_body(V2) :- - preamble(V2), not pomegranate(V2), not asset(V2).
coptic_church(V2, V3) :- - restaurant_dining_area(V3), alarm(V2).
coptic_church(V2, V3) :- launching(V2), 
    - restaurant_dining_area(V3), not adult_body(V2).
- deadwood(V0) :- coptic_church(V0, V1).
- rocks(V0, V1) :- coptic_church(V0, V1).
- rocks(V1, V1) :- stagehand(V1, V2).
asset(V2) :- stagehand(V1, V2), not print_on(V2), - adult_body(V2).
- rocks(V2, V1) :- alarm(V2), stagehand(V1, V2), not cell(V1, V2).
- slope(V2, V2) :- - deadwood(V2), not -preamble(V2), not -dividend_warrant(V2).
spearhead(V2) :- dolphin(V0, V1), - cow_juice(V1, V2).
ruining_appetite(V0, V2) :- larid(V0, V2), not -snoop(V0, V2), feeling_good(V2, V3).
\end{verbatim}
\\ \hline
\textbf{\textsc{Query:}} \texttt{coptic\_church("Savannah", "Matthew").}
\\ \hline
\textbf{\textsc{Label:}} \texttt{False} \\ \hline
\textbf{\textsc{Response of LLMs:}} \\

\\
\textit{o4-mini:} \\
\definecolor{darkgreen}{rgb}{0,0.5,0}%
... Derives \texttt{arsenic("Savannah","Matthew")} and \texttt{restaurant\_dining\_area("David")}.
Key correct derivations:
Rule 9 leads to \textcolor{darkgreen}{\texttt{-preamble("Matthew")}}. 
Then, Rule 10 leads to \textcolor{darkgreen}{\texttt{rocks("Matthew","Matthew")}}. 
... Derives \texttt{-launching("Scott")}.
Rule 2, with \texttt{-launching("Scott")} and \texttt{rocks("Matthew","Matthew")}, leads to \textcolor{darkgreen}{\texttt{deadwood("Scott")}}. 
... Finally, Rule 1 derives \texttt{-coptic\_church("Savannah","Matthew")}.
The query is \textcolor{darkgreen}{False}. \\

\\
\textit{deepseek-r1:} \\
... Derived \texttt{arsenic("Savannah","Matthew")} and \texttt{-launching("Scott")}.
However, the reasoning faltered:
It attempted to derive \texttt{rocks(V0,V1)} by first targeting \textcolor{red}{\texttt{-preamble("Scott")}}, which failed as its premise \texttt{arsenic(V0,"Scott")} was underivable (due to no fact \texttt{-wall(V0,"Scott")}).
This led to \textcolor{red}{\texttt{rocks("Scott","Scott")} (or any relevant \texttt{rocks})} not being derivable.
Consequently, \textcolor{red}{\texttt{deadwood("Scott")}} could not be derived via Rule 2.
... Concluding that \texttt{-coptic\_church("Savannah","Matthew")} was not derivable.
Resulting in the \textcolor{red}{Unknown}. \\

\\
\textit{qwen3-14b:} \\
... However, when trying to derive \texttt{-coptic\_church("Savannah","Matthew")} via Rule 1 (which requires \texttt{deadwood(V2)}), the reasoning faltered:
It concluded that \textcolor{red}{\texttt{deadwood(V2)} was not derivable} because its prerequisite \texttt{rocks(V0,V1)} (needed for Rule 2 to derive \texttt{deadwood}) could not be established.
... This led to the inability to derive \texttt{-coptic\_church("Savannah","Matthew")}.
As positive derivations for \texttt{coptic\_church("Savannah","Matthew")} also failed, the final conclusion was \textcolor{red}{Unknown}. \\

\\ \bottomrule
\end{longtable}
\twocolumn
\onecolumn 
\small
\begin{longtable}{p{\textwidth}}
\caption{Case study for answer set verification}\label{tab:case_study_ASV} \\
\toprule 
\textbf{\textsc{Input ASP Program:}} \\ \hline
\begin{verbatim}
- americana("Andrew", "Angela").
- baa_lamb("Angela", "David").
- consumer_goods("David").
- ledger.
- oriental("Lauren", "Andrew").
- out("Lauren", "Andrew").
- pledge("Lauren").
- write_letter("Andrew", "Angela").
grassy("David").
nestorian_church("Angela", "David").
pickerel("Lauren").
pubes("Andrew", "David").
shrub("Lauren", "Andrew").
smart_bomb("Lauren", "Angela").
smoke.
- cost_cutting(V0, V2) | - coronal_suture(V0) :- - oriental(V0, V1), smart_bomb(V0, V2), 
    - write_letter(V1, V2), - baa_lamb(V2, V3), not -nestorian_church(V2, V3).
- depth(V3, V3) :- - fraternization(V3).
- france(V0, V1) :- not fraternization(V3), - adults(V0, V1), - sacrum(V0, V3).
- france(V0, V3) :- - adults(V0, V1), - sacrum(V0, V3).
- new_zealand(V2, V2) :- cost_cutting(V0, V2), smoke.
- winter_sport(V0, V3) :- - out(V0, V1), pubes(V1, V3).
back(V0, V3) :- adults(V0, V1), new_zealand(V1, V3).
consumer_goods(V3) :- - adults(V0, V1), - new_zealand(V1, V3), not -shrub(V0, V1).
coronal_suture(V0) :- star_in_moview(V0, V2).
depth(V0, V2) | adults(V0, V2) :- - cost_cutting(V0, V2), not star_in_moview(V0, V2).
france(V0, V2) :- fraternization(V3), depth(V0, V2), not -sacrum(V0, V3).
fraternization(V3) :- grassy(V3), not consumer_goods(V3).
ledger(V0, V2) :- not -star_in_moview(V0, V2), - france(V0, V2).
new_zealand(V2, V0) | - star_in_moview(V0, V2) :- - cost_cutting(V0, V2), 
    not coronal_suture(V0), not ledger.
sacrum(V0, V3) :- back(V0, V3), not coronal_suture(V0), 
    not consumer_goods(V3), not pledge(V0), not -pickerel(V0).
scantling(V1) :- - americana(V1, V2).
star_in_moview(V0, V2) :- - back(V0, V3), cost_cutting(V0, V2), not france(V0, V2).
\end{verbatim}
\\ \hline
\textbf{\textsc{Query Answer Set:}} 
\begin{verbatim}
{ -americana("Andrew","Angela"),
-baa_lamb("Angela","David"),
-consumer_goods("David"),
-cost_cutting("Lauren","Angela"),
-ledger,
-oriental("Lauren","Andrew"),
-out("Lauren","Andrew"),
-pledge("Lauren"),
-winter_sport("Lauren","David"),
-write_letter("Andrew","Angela"),
depth("Lauren","Angela"),
france("Lauren","Angela"),
fraternization("David"),
grassy("David"),
nestorian_church("Angela","David"),
new_zealand("Angela","Lauren"),
pickerel("Lauren"),
pubes("Andrew","David"),
scantling("Andrew"),
shrub("Lauren","Andrew"),
smart_bomb("Lauren","Angela"),
smoke }
\end{verbatim}
\\ \hline
\textbf{\textsc{Label:}} \texttt{Is Answer Set} \\ \hline
\textbf{\textsc{Response of LLMs:}} \\
\\
\textit{o3-mini:} \\
``The original rule 6 is \textcolor{red}{\texttt{winter\_sport(V0, V3) :- -out(V0, V1), pubes(V1, V3)}}.
...Hence the rule forces the head: \textcolor{red}{\texttt{winter\_sport("Lauren", "David")}} to be true in every model of the reduct.
...Now look at the candidate set A. It does not contain \texttt{winter\_sport("Lauren", "David")}. Instead, it contains \texttt{-winter\_sport("Lauren", "David")}.''
\textcolor{red}{``Final Answer: No, the candidate answer set is not a valid answer set.''} \\\\
\\
\textit{o4-mini:} \\
\definecolor{darkgreen}{rgb}{0,0.5,0}%
... Rule 6: from {-out("Lauren","Andrew")} and {pubes("Andrew","David")} we derive \textcolor{darkgreen}{\texttt{-winter\_sport("Lauren","David")} $\in$ \texttt{A}}. ...
\textcolor{darkgreen}{Final Answer: Yes. The given candidate set is a valid (stable) answer set of the program.} \\
\\
\textit{gemini-2.5-flash-thinking:} \\
... * **Rule 2:** \texttt{- depth(V3, V3) :- - fraternization(V3).}
* Body \texttt{-fraternization(V3)}. No literal \texttt{-fraternization(...)} is in A. No ground instance body is true in A. \textcolor{red}{No rule added to $P^A$ based on A}. ... \\

\bottomrule
\end{longtable}
\twocolumn
\onecolumn 
\small
\begin{longtable}{p{\textwidth}}
\caption{Case study for answer set computation. Blue highlighting in output of deepseek-r1 indicates literals missed by o4-mini. Orange highlighting in output of deepseek-r1 indicates literals missed by qwen3-14b. Red highlighting indicates errors in a reasoning of an LLM or its incorrect output.}\label{tab:case_study_ASC} \\
\toprule 
\textbf{\textsc{Input ASP Program:}} \\ \hline
\begin{verbatim}
- P17("Christopher").
- P18("John", "Charles", "Lucas").
- P19("John", "Charles", "Lucas").
- P20("John", "Charles").
- P21("Barbara", "John", "Lucas").
- P24("Christopher", "John", "Charles").
- P28("Christopher", "John", "Charles").
- P30("Barbara", "Christopher", "John").
P12("Barbara", "Christopher").
P13("Barbara", "John", "Charles").
P14("Barbara", "John", "Lucas").
P15("Christopher", "John", "Lucas").
P16("Christopher", "Charles", "Lucas").
P22("Barbara", "John", "Lucas").
P23("Barbara", "John", "Lucas").
P25("Christopher", "Charles").
P26.
P31("Barbara", "Christopher", "Lucas").
P32("Barbara", "Christopher", "Lucas").
- P0(V1, V3) :- P2(V1, V3, V4), not P5(V1, V3, V4), not -P16(V1, V3, V4).
- P1(V1) :- P0(V1, V3), P5(V1, V3, V4), P25(V1, V3).
- P11(V1, V3, V4) :- - P0(V1, V3), - P4(V0, V1, V3), - P6(V0, V1, V4).
- P12(V0, V1) :- P3(V0, V2, V4), P11(V1, V3, V4).
- P14(V0, V2, V4) :- P0(V1, V3), P3(V0, V2, V4), P6(V0, V1, V4).
- P29(V2, V4) :- - P30(V0, V1, V2), P31(V0, V1, V4), not -P32(V0, V1, V4).
- P3(V0, V2, V4) :- P12(V0, V1), - P20(V2, V3), - P21(V0, V2, V4).
- P4(V0, V1, V3) :- P13(V0, V2, V3), P14(V0, V2, V4), P15(V1, V2, V4).
- P5(V1, V3, V4) :- - P7(V2, V4), P10(V0, V1, V4), - P24(V1, V2, V3).
- P6(V1, V1, V1) :- P1(V1), not P8(V1), not P17(V1).
- P7(V2, V4) :- - P3(V0, V2, V4), not -P22(V0, V2, V4), not -P23(V0, V2, V4).
- P8(V0) :- - P3(V0, V2, V4), not -P22(V0, V2, V4), not -P23(V0, V2, V4).
- P9(V2, V0, V4) :- - P3(V0, V2, V4), not -P22(V0, V2, V4), not -P23(V0, V2, V4).
P1(V1) :- P2(V1, V3, V4), not P5(V1, V3, V4), not -P16(V1, V3, V4).
P10(V4, V4, V4) :- - P9(V2, V3, V4), not P18(V2, V3, V4), not P19(V2, V3, V4).
P11(V1, V3, V4) :- P8(V1), P9(V2, V3, V4), - P10(V0, V1, V4).
P11(V1, V4, V4) :- P3(V0, V2, V4), not -P6(V0, V1, V4), - P10(V0, V1, V4).
P18(V3, V3, V3) :- P0(V1, V3), not P1(V1), P8(V1).
P19(V1, V1, V1) :- - P1(V1), not -P26.
P2(V1, V3, V4) :- P12(V0, V1), - P20(V2, V3), - P21(V0, V2, V4).
P27(V1, V2, V3) :- - P28(V1, V2, V3).
P6(V0, V1, V4) :- P0(V1, V3), P3(V0, V2, V4), P11(V1, V3, V4).
\end{verbatim}
\\ \hline
\textbf{\textsc{Response of LLMs:}} \\
\\
\textit{deepseek-r1:} \\
Generated Answer Set: { -P17("Christopher"), -P18("John", "Charles", "Lucas"), -P19("John", "Charles", "Lucas"), -P20("John", "Charles"), -P21("Barbara", "John", "Lucas"), -P24("Christopher", "John", "Charles"), -P28("Christopher", "John", "Charles"), -P30("Barbara", "Christopher", "John"), P12("Barbara", "Christopher"), P13("Barbara", "John", "Charles"), P14("Barbara", "John", "Lucas"), P15("Christopher", "John", "Lucas"), \textcolor{orange}{P16("Christopher", "Charles", "Lucas")}, P22("Barbara", "John", "Lucas"), P23("Barbara", "John", "Lucas"), P25("Christopher", "Charles"), P26, P31("Barbara", "Christopher", "Lucas"), P32("Barbara", "Christopher", "Lucas"), \textcolor{orange}{P2("Christopher", "Charles", "Lucas")}, \textcolor{blue}{P1("Christopher")}, P27("Christopher", "John", "Charles"), \textcolor{blue}{-P0("Christopher", "Charles")}, \textcolor{blue}{-P6("Christopher", "Christopher", "Christopher")}, -P3("Barbara", "John", "Lucas"), -P4("Barbara", "Christopher", "Charles"), -P29("John", "Lucas"), -P7("John", "Lucas"), -P8("Barbara"), -P9("John", "Barbara", "Lucas"), \textcolor{orange}{P10("Lucas", "Lucas", "Lucas")} } \\\\
\\
\textit{o4-mini:} \
Here is one candidate stable model. In what follows I list all ground literals that end up true (including both the original facts and everything forced by the rules under default negation): { –P17("Christopher"), –P18("John","Charles","Lucas"), –P19("John","Charles","Lucas"), –P20("John","Charles"), –P21("Barbara","John","Lucas"), –P24("Christopher","John","Charles"), –P28("Christopher","John","Charles"), –P30("Barbara","Christopher","John"), P12("Barbara","Christopher"), P13("Barbara","John","Charles"), P14("Barbara","John","Lucas"), P15("Christopher","John","Lucas"), P16("Christopher","Charles","Lucas"), P22("Barbara","John","Lucas"), P23("Barbara","John","Lucas"), P25("Christopher","Charles"), P26, P31("Barbara","Christopher","Lucas"), P32("Barbara","Christopher","Lucas"), P2("Christopher","Charles","Lucas"), (from rule 20) –P3("Barbara","John","Lucas"), (from rule 7) –P4("Barbara","Christopher","Charles"), (from rule 8) –P7("John","Lucas"), (from rule 11) –P8("Barbara"), (from rule 12) –P9("John","Barbara","Lucas"), (from rule 13) –P29("John","Lucas"), (from rule 6) P10("Lucas","Lucas","Lucas"), (from rule 15) P27("Christopher","John","Charles") (from rule 21) } One can check that (1) no atom and its strong negation both occur, and (2) under the Gelfond–Lifschitz reduct w.r.t. this set \textcolor{red}{all and only these literals are derivable}.\\
\\
\textit{qwen3-14b:} \\
"...rule 8: - P4(V0, V1, V3) :- P13(V0, V2, V3), P14(V0, V2, V4), P15(V1, V2, V4). ... so \textcolor{red}{-P4("Barbara", "John", "Charles")} is derived." 
... 
"...rule 15: P10(V4, V4, V4) :- - P9(V2, V3, V4), not P18(V2, V3, V4), not P19(V2, V3, V4). ... P18("John", "Barbara", "Lucas") is not mentioned in the facts. \textcolor{red}{Therefore, unless there is a rule that derives it, we can't assume it is true or false.} ... Therefore, maybe this rule doesn't apply here." \
... 
Final Predicted Answer Set by qwen3-14b (Incorrect): 
{ -P17("Christopher"), -P18("John", "Charles", "Lucas"), -P19("John", "Charles", "Lucas"), -P20("John", "Charles"), -P21("Barbara", "John", "Lucas"), -P24("Christopher", "John", "Charles"), -P28("Christopher", "John", "Charles"), -P30("Barbara", "Christopher", "John"), P12("Barbara", "Christopher"), P13("Barbara", "John", "Charles"), P14("Barbara", "John", "Lucas"), P15("Christopher", "John", "Lucas"), P22("Barbara", "John", "Lucas"), P23("Barbara", "John", "Lucas"), P25("Christopher", "Charles"), P26, P31("Barbara", "Christopher", "Lucas"), P32("Barbara", "Christopher", "Lucas"), P1("Christopher"), -P0("Christopher", "Charles"), -P3("Barbara", "John", "Lucas"), -P8("Barbara"), -P9("John", "Barbara", "Lucas"), P27("Christopher", "John", "Charles"), \textcolor{red}{-P4("Barbara", "John", "Charles")}, -P7("John", "Lucas"), -P29("John", "Lucas"), -P6("Christopher", "Christopher", "Christopher")} \\

\bottomrule
\end{longtable}
\twocolumn

\end{document}